\documentclass[letterpaper]{article} % DO NOT CHANGE THIS
\usepackage{aaai2027}  % DO NOT CHANGE THIS
\usepackage[hyphens]{url}  % DO NOT CHANGE THIS
\usepackage{graphicx} % DO NOT CHANGE THIS
\usepackage{natbib}  % DO NOT CHANGE THIS AND DO NOT ADD ANY OPTIONS TO IT
\usepackage{caption} % DO NOT CHANGE THIS AND DO NOT ADD ANY OPTIONS TO IT
\usepackage{algorithm}
\usepackage{algorithmic}
\usepackage{graphicx}
\usepackage{makecell}
\usepackage{amsmath}
\usepackage{booktabs} % For formal tables
\usepackage{multirow}
\usepackage{amssymb}

\usepackage{newfloat}
\usepackage{listings}
\DeclareCaptionStyle{ruled}{labelfont=normalfont,labelsep=colon,strut=off} % DO NOT CHANGE THIS
\floatstyle{ruled}
\newfloat{listing}{tb}{lst}{}
\floatname{listing}{Listing}

\usepackage{booktabs}

\title{GDLAM: Group-Disentangled Latent Action Model for Highly Disentangled Embodied Pretraining}
\author {
Jiarui Yang\textsuperscript{\rm 1,\rm 2}, Jiawei Li\textsuperscript{\rm 2}, Jiale Zhang\textsuperscript{\rm 2}, Hang Guo\textsuperscript{\rm 4}, Wen Huang\textsuperscript{\rm 2}, Maowei Hu \textsuperscript{\rm 2}, Tao Dai \textsuperscript{\rm 3}, Shu-Tao Xia\textsuperscript{\rm 2}
}
\affiliations {
    \textsuperscript{\rm 1} Nankai University 
    \textsuperscript{\rm 2} Tsinghua University
    \textsuperscript{\rm 3} Shenzhen University \\
    \textsuperscript{\rm 4} Swiss Federal Institute of Technology in Lausanne (EPFL) 
}

\begin{document}

\maketitle

\begin{abstract}
% 潜在动作模型（LAMs）通过自监督未来预测从无动作标注视频中学习动作相关表征，为具身智能预训练提供了可扩展方案。然而，现有 LAMs 通常将相机运动、物体运动、交互事件等不同因素混合编码为单一潜在向量，导致表征缺乏结构化语义，限制了世界模型控制能力和 VLA 策略的泛化性能。我们提出分组解耦潜在动作模型（GDLAM），该模型的编码通过设计被按构造分解为N个组，每个组具有独立的变分瓶颈和空间门控路由路径，并通过一组信息几何目标进行训练——互斥性、组稀疏性与门控稀疏性、静态-动态正交性——使得各组之间因果独立而非仅仅不相关。在定量评估中，对任意单一组进行干预仅改变该组而保持其他组不变（呈现强对角线的跨组响应），且GDLAM在无监督解耦指标——模块化性、互信息差距（MIG）和离散化互信息（DCI）——上较强大的非结构化LAMs取得显著提升。作为解耦编码可作为可复用预训练资源的佐证，我们进一步将其迁移至两个下游场景 1:世界模型：基于GDLAM的世界模型相对于SOTA基线在Rollout具有更优越的性能和动作遵循；2:VLA：在多个基准和真机上，预训练相比之前方法大幅提升了任务的成功率。

Latent action models (LAMs) learn action-related representations from action-free videos via self-supervised future prediction, offering a scalable paradigm for embodied intelligence pretraining. However, existing LAMs collapse heterogeneous sources of visual change—including camera motion, object dynamics, and interaction events—into a single latent vector, resulting in entangled representations with limited semantic structure and consequently restricting world model controllability and VLA policy generalization. We introduce the \textbf{Group-Disentangled Latent Action Model (GDLAM)}, a latent action model whose code is factorized by construction into $N$ groups, each with an independent variational bottleneck and a spatially gated routing pathway, and trained with a set of information-geometric objectives---mutual exclusivity, group and gate sparsity, and static--dynamic orthogonality---that make the groups mutually causally distinct rather than merely decorrelated. Quantitatively, intervening on any single group changes only that group and leaves the others intact (a strongly diagonal cross-group response), and GDLAM improves label-free disentanglement metrics---Modularity, MIG, and DCI---by wide margins over a strong unstructured LAM. Notably, this factorization is not at the expense of action information: across three mutual-information estimators and a linear probe, the grouped code is more informative than monolithic baselines both in- and out-of-distribution. As supporting evidence that the disentangled code is a reusable pretraining currency, we further transfer it to two downstream regimes: (1) \textbf{World Modeling}: World models pretrained with GDLAM achieve superior rollout fidelity and action-following capability compared with SOTA baselines. (2) \textbf{VLA Policies}: Pretraining with GDLAM substantially improves task success rates over previous methods across multiple simulation benchmarks and real-world robotic manipulation tasks.
\end{abstract}

\section{Introduction}

\begin{figure}[t]
\centering
\includegraphics[width=0.99\columnwidth]{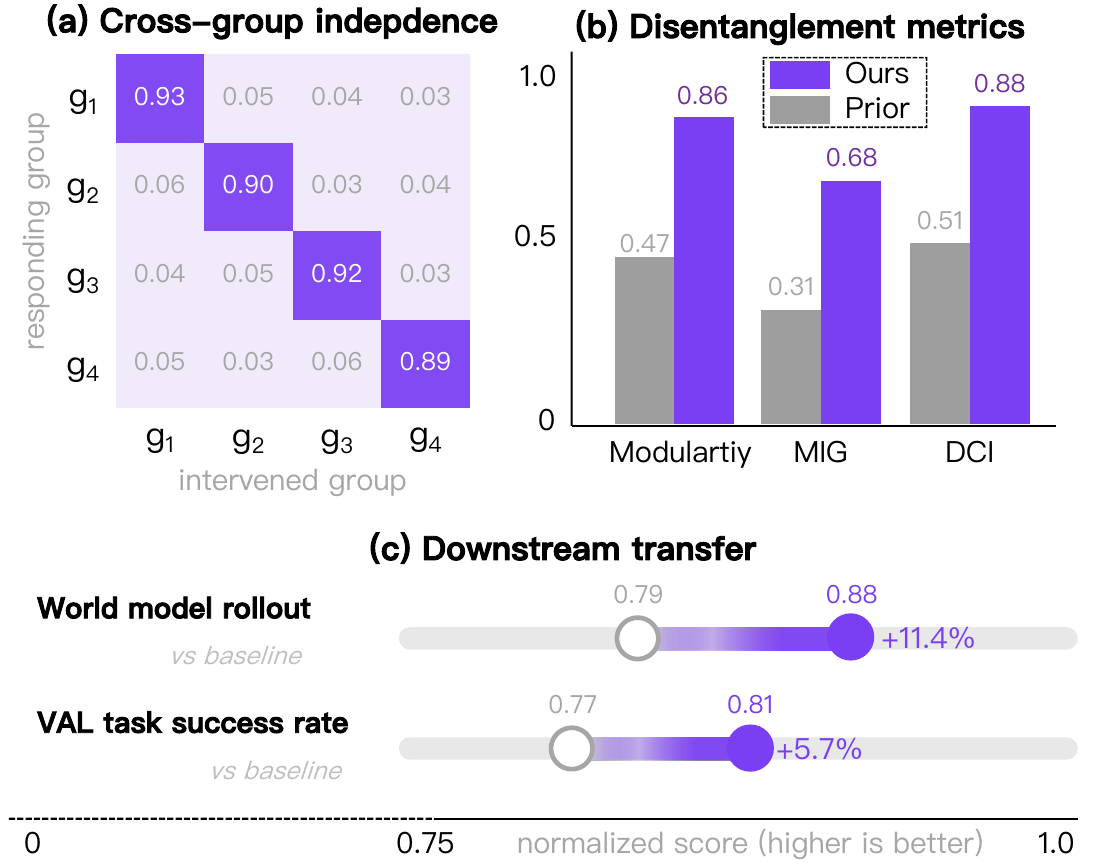}
\caption{GDLAM learns disentangled latent actions. (a) Cross-group response matrix under single-group intervention. A diagonal response indicates strong group independence. (b) Higher unsupervised disentanglement scores than a prior LAM. (c) Better downstream transfer.}
\label{fig:teaser}
\end{figure}

% 从原始、无动作标注的视频中学习动作，是突破小规模遥操作机器人数据限制、实现具身智能规模化发展的关键途径之一。潜在动作模型通过一个逆向–前向建模框架实现这一思想：编码器从连续两帧观测中推断潜在动作，而解码器则在该潜在动作条件下预测未来帧。由于其唯一监督信号来源于自监督的未来预测任务，LAMs 能够利用互联网规模以及第一视角（egocentric）视频进行训练。同时，所推断出的潜在动作逐渐成为一种具有吸引力的预训练媒介——它提供了紧凑且可迁移的表示信号，可用于初始化下游策略模型以及生成式世界模型。

Learning to act from raw, actionless video is a central lever for scaling embodied intelligence beyond the small pools of teleoperated, action-labeled robot data \citep{gao2026dreamdojo}. Latent action models (LAMs) operationalize this idea through an inverse--forward pair: an encoder infers a latent action from a pair of consecutive frames, and a decoder predicts the future frame conditioned on that action \citep{lapa}. Because the only supervision is self-supervised future prediction, LAMs can be trained on internet-scale and egocentric video \citep{egodex}, and the inferred latent actions have become an attractive pretraining currency---compact, transferable signals that can seed downstream policies and world models \citep{cen2025worldvla}.

% 本文的核心前提是：潜在动作模型（LAM）不应仅依据其未来重建能力进行评价，更应关注其潜在编码作为预训练表示的有效性。从这一视角来看，现有 LAM 中占主导地位的设计选择——单一、无结构的潜在向量——实际上构成了其主要局限。经验上，该单一潜在向量会将视觉变化背后的多种异质因素纠缠在一起：包括全局相机或基座运动、机械臂/手部的关节运动、离散的接触与抓取事件，以及缓慢变化的场景外观漂移等因素，均被叠加编码于同一个潜在空间中。这种纠缠结构在预训练阶段带来了两个相互叠加的代价。首先，它限制了表示的可解释性与可干预性。由于不同变化因素无法被独立分离和操控，下游模型必须从头开始隐式地重新学习潜在因素之间的分解关系，本质上重复完成对异质动力学因素的解耦过程。其次，当这种纠缠表示用于条件化世界模型时，模型缺乏能够分别控制底层动力学过程的因子化控制变量，导致条件生成的未来轨迹预测更加不准确，同时也降低了模型的可控性。类似地，当该表示用于初始化视觉-语言-动作（VLA）策略模型时，由于潜在维度与稳定语义因素之间缺少明确对应关系，语言指令难以可靠地映射到具体动作因素，从而降低了样本效率，并限制了策略的组合泛化能力。

The premise of this paper is that a LAM should be evaluated not only by future reconstruction quality, but also by the usefulness of its latent code as a representation. From this perspective, the prevailing design of prior LAMs---a single unstructured latent vector---is a limitation \citep{wang2024disentangled}. Such a code entangles diverse causes of visual change, including camera/base motion, articulated manipulation, contact events, and scene dynamics. This entanglement harms downstream learning in two ways: it prevents explicit factor manipulation, forcing downstream models to relearn latent factorization, and it provides no factorized control variables for world models or VLA policies, reducing rollout controllability, language-action grounding, and compositional generalization \citep{genie, rt2}.

% 因此，我们认为，解耦性应当成为预训练表征本身的一项核心属性，而非下游任务再去学习的附加能力。基于这一观点，我们提出了组解耦潜在动作模型GDLAM。GDLAM首先利用冻结的视觉基础模型Tokenizer构建稳定的感知空间，并在模型结构层面将潜在动作显式划分为N个组。每个组都拥有一个可学习的前缀查询、一个独立的变分瓶颈以及一个超球面投影。随后，通过一个具有空间门控路由的预测器，每个组沿着专属的交叉注意力路径注入模型；其中，零初始化的门控参数使各组能够逐渐专门化到不同的空间区域和不同类型的视觉变化模式。在此基础上，我们进一步引入四种信息几何约束——互斥性、组稀疏性、门控稀疏性以及静态–动态正交性——共同推动不同组学习到因果意义上的相互独立，而不仅仅是偶然实现统计上的去相关。这些组作为一组可干预、彼此独立的潜在基底，其语义是由干预后的行为表现所决定，而非预先假设的身份定义。如图~\ref{fig:teaser}(a)所示，所得模型的跨组响应矩阵呈现出显著的对角化结构：对某一组进行扰动几乎不会影响其他各组，表明各组之间具有良好的因果独立性。与此同时，图~\ref{fig:teaser}(b)进一步表明，相较于采用无结构潜在表示的传统LAM，GDLAM在无标签解耦指标上取得了显著提升。

We argue that disentanglement should be a first-class property of the pretraining representation itself and propose the \textbf{Group-Disentangled Latent Action Model (GDLAM)}. GDLAM fixes a stable perceptual space with a frozen vision-foundation-model (VFM) tokenizer \citep{dinov2} and factorizes the latent action by construction into $N$ groups: each group owns a learnable prefix query, an independent variational bottleneck, and a hyperspherical projection, and is injected by a spatially gated routing predictor through a dedicated cross-attention pathway whose zero-initialized gate lets groups specialize to distinct regions and modes of change. Four information-geometric objectives---mutual exclusivity, group sparsity, gate sparsity, and static--dynamic orthogonality---then push the groups to be mutually causally distinct rather than decorrelated by luck. The groups emerge as an intervenable, mutually independent basis, characterized by how they behave under intervention rather than by an assumed identity. Fig.~\ref{fig:teaser}(a) shows the resulting cross-group response matrix is strongly diagonal---perturbing one group leaves the others essentially unchanged---and Fig.~\ref{fig:teaser}(b) shows large gains on label-free disentanglement metrics over an unstructured LAM. This grouping comes at no cost to action information: under three mutual-information estimators and a linear probe, the grouped code stays more informative than monolithic baselines both in- and out-of-distribution (OOD).

% 我们进一步在世界模型和 VLA 的预训练–后训练范式中验证 GDLAM 所学习的解耦潜在动作表示的迁移能力。首先，我们从大规模无动作视频数据中提取潜在动作标签，并分别用于世界模型和 VLA 的预训练：前者将潜在动作作为条件变量进行未来状态生成，后者学习从当前观测预测结构化潜在动作。随后，两类模型均通过真实动作数据进行后训练，以适配下游控制任务。我们采用包含 16 项指标的综合基准以及真实世界 OOD 测试评估世界模型 Rollout 质量，并在多样化仿真环境和真实机器人平台上评估 VLA 的任务成功率。进一步的大规模消融实验验证了 GDLAM 分组解耦机制的有效性及各组件的必要性。我们的贡献如下：

We further validate the transferability of the disentangled latent action representations learned by GDLAM under the pretraining–post-training paradigms of both world models and VLA policies. Specifically, we first extract latent action labels from large-scale actionless video data and leverage them for pretraining in two downstream settings. For world model pretraining, the extracted latent actions serve as conditional variables to guide future state generation \citep{gao2026dreamdojo}. For VLA pretraining, the model learns to predict structured latent actions from current observations, thereby acquiring a semantically organized action representation \citep{mvplam}. Subsequently, both models are post-trained on datasets with real action supervision to adapt to downstream control tasks. For world models, we evaluate video rollouts using a comprehensive benchmark with 16 metrics, together with real-world OOD generalization tests. For VLA policies, we assess task success rates across diverse simulation environments and real robotic platforms. Furthermore, extensive ablation studies demonstrate the effectiveness of GDLAM's group-wise disentanglement mechanism and verify the contribution of each proposed component. Our contributions are summarized as follows:

% 我们提出 \textbf{GDLAM}，通过组化潜在编码、独立变分瓶颈和信息几何约束，实现潜在动作的因果解耦。我们从因果角度验证解耦效果：单组干预产生强对角化响应矩阵；GDLAM 在 Modularity、MIG、DCI 上显著优于现有 LAM，并在分布内外保持更丰富的动作信息。我们证明了解耦编码具有预训练复用价值：迁移至组条件世界模型和结构化 VLA 策略后，在仿真与真实机器人任务中均提升性能。

\begin{itemize}
    \item We propose \textbf{GDLAM}, which achieves causal disentanglement of latent actions through grouped latent factorization, independent variational bottlenecks, and information-geometric constraints.
    \item We validate disentanglement from a causal perspective: single-group intervention produces a strongly diagonal cross-group response matrix; GDLAM substantially outperforms existing LAMs on Modularity, MIG, and DCI, while preserving more informative action representations both in- and OOD.
    \item We demonstrate the pretraining reusability of the disentangled representation: transferring it to world models and structured VLA policies consistently improves performance on both simulation benchmarks and real-world robotic tasks.
\end{itemize}

\begin{figure*}[t]
\centering
\includegraphics[width=0.98\textwidth]{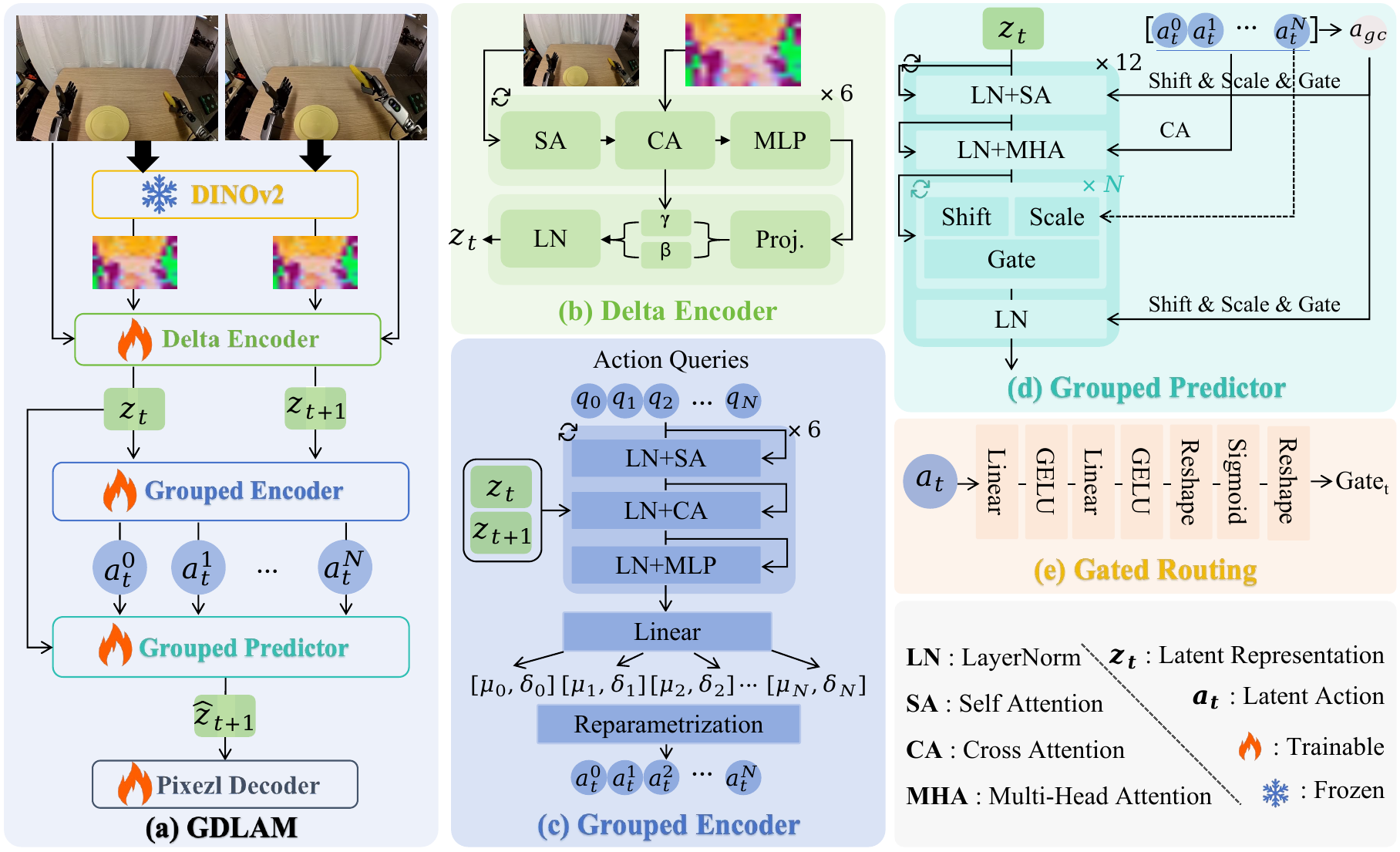}
\caption{\textbf{Overview of GDLAM.} \textbf{(a)} Main pipeline: a frozen DINOv2 tokenizer and a trainable Delta Encoder map two frames to latents $z_t,z_{t+1}$; a Grouped Encoder reads them into $N$ latent actions $a_t^1,\dots,a_t^N$; a Grouped Predictor transforms $z_t$ conditioned on the actions into the predicted future $\hat{z}_{t+1}$. (b--e) Building blocks: the Delta Encoder, the Grouped Encoder with per-group variational bottlenecks, the Grouped Predictor, and the gated routing that forms each group's spatial gate $\mathrm{Gate}_t$.}
\label{fig:overview}
\end{figure*}

\section{Related Work}

\paragraph{Latent action models.} LAMs infer action-like latents by self-supervised future-frame prediction. Genie \citep{genie} learns discrete latent actions with a VQ codebook to drive an interactive video generator; LAPA \citep{lapa} uses latent actions to pretrain robot policies from actionless video. These models treat the latent action as a single unstructured (often discrete) code \citep{gao2026dreamdojo, adaworld}, so any downstream factorization must be relearned. Recent methods, such as UniVLA \citep{univla} and MVP-LAM \citep{mvplam}, further replace pixel-space tokenizers with latent-space representations, enabling semantically richer and more stable latent representations. However, these approaches still represent latent actions as a monolithic code, leaving the underlying factors of visual change entangled and requiring downstream models to implicitly disentangle them.

\paragraph{Disentangled representation learning} aims to learn latent representations in which different generative factors are encoded independently, such that each latent dimension or subspace corresponds to a specific semantic factor. VAE-based approaches promote disentanglement through a single global bottleneck: $\beta$-VAE \citep{burgess2018understanding} increases the weight of the KL regularization term to encourage factorized posteriors, while FactorVAE \citep{kim2018disentangling} and $\beta$-TCVAE \citep{chen2018isolating} explicitly constrain the total correlation among latent variables. InfoGAN \citep{chen2016infogan} improves interpretability by maximizing the mutual information between latent codes and generated outputs. Beyond single-bottleneck formulations, another line of work introduces explicit structural decomposition of the latent space. For example, capsule networks \citep{sabour2017dynamic} and slot-based models \citep{locatello2020object} partition representations into multiple sub-vectors and assign different generative factors to corresponding subspaces, providing a more structured and factorized representation.

\section{Method}

\subsection{Problem Formulation}
Let $\{x_t\}$ be frames of actionless video. A LAM factorizes future prediction as:
\begin{equation}
    p(x_{t+1}\mid x_t)=\int p(x_{t+1}\mid x_t,\mathbf{a}_t)\,q(\mathbf{a}_t\mid x_t,x_{t+1})\,d\mathbf{a}_t
\end{equation} where $\mathbf{a}_t$ is a latent action inferred by the encoder $q$ and consumed by the predictor $p$ (Fig.~\ref{fig:overview}(a)). The visual change $\delta_t = x_{t+1}\!\ominus\!x_t$ is generated by a small number of heterogeneous causes---acting on different image regions and time scales---that a monolithic $\mathbf{a}_t$ marginalizes over, so no downstream module can intervene on a single cause without disturbing the others. Our goal is to recover a code whose parts are mutually intervenable regardless of what each part ends up encoding.

GDLAM instead posits a generative factorization $\mathbf{a}_t=\{a_t^{(g)}\}_{g=1}^{N}$ into $N$ groups, and enforces three principles: (i) capacity isolation---each group owns an independent stochastic bottleneck so groups cannot silently share capacity; (ii) spatial specialization---each group is injected through a gated pathway that can claim the image region it explains; and (iii) causal distinctness---explicit objectives push groups to be mutually exclusive and sparse, so an intervention on $a_t^{(g)}$ changes only that group. The only semantic prior we impose is a coarse time-scale split: one designated group is encouraged to carry slow, static appearance while the remaining groups carry change. These principles make the code a reusable, factorized pretraining representation whose parts are individually intervenable.

\subsection{Tokenizer as a Stable Perceptual Space}
Each frame is encoded by a frozen vision foundation model whose class and register tokens are discarded, yielding a grid of patch tokens $u_t$ that we treat as a fixed perceptual manifold. A lightweight trainable Delta Encoder refines these tokens (Fig.~\ref{fig:overview}(b)): each layer applies self-attention (SA) over the patch grid, then cross-attention (CA) whose queries are the current tokens and whose keys/values are the frozen VFM features, followed by an MLP (all with LayerScale residuals). The encoder fuses its refined stream back onto the frozen features $u_t$ through a feature-wise linear modulation (FiLM \citep{dumoulin2018feature}): a zero-initialized projection predicts per-channel $(\gamma,\beta)$ and the fused latent representation is $z_t=\mathrm{LN}\big(u_t\odot(1+\gamma)+\beta\big)$, so the Delta Encoder starts as the identity on the frozen manifold and only injects change as it is learned. A latent compressor (attention$+$Conv) reduces each token of $z_t$ to a compact code of dimension $d{=}32$ that is $\ell_2$-normalized per token onto the hypersphere $\mathbb{S}^{d-1}$. Applying this to both frames yields the pair $z_t,z_{t+1}$.

\subsection{Grouped Action Encoder with Isolated Bottlenecks}
The encoder maps the frame pair $(z_t,z_{t+1})$ to $N$ latent actions ($\tau{=}1$ token per group), indexed $a_t^1,\dots,a_t^N$ (Fig.~\ref{fig:overview}c). The two frames' tokens are each augmented with a spatial (sinusoidal) position embedding and a learned temporal embedding, then concatenated to form the key/value memory. A set of $N$ learnable action queries $q_1,\dots,q_N$ attends to this memory through a $6$-layer perceiver-style transformer (LN$+$SA among queries, LN$+$CA into the memory, LN$+$MLP). Each group's query is seeded with a learnable group prefix embedding, giving it a distinct read-out bias over the change representation. After the shared trunk, each group $g$ passes through its own variational bottleneck (an independent linear head) producing $[\mu_g,\delta_g]$, where $\delta_g$ parameterizes the log-variance; we clamp $\delta_g$ for stability, reparameterize and renormalize the resulting $d$-dimensional latent onto the hypersphere to obtain $a_t^{g}$. The isolated bottlenecks force each group to pay its own rate--distortion cost, such that a factor is encoded only if it provides a genuine reduction in prediction error.

\begin{table}[t]
\centering
\caption{Label-free disentanglement scores (higher is better for all three metrics).}
\label{tab:disent}
\begin{tabular}{lccc}
\toprule
Model & Modularity$\uparrow$ & MIG$\uparrow$ & DCI$\uparrow$ \\
\midrule
UniVLA    & 0.47 & 0.31 & 0.51 \\
LAPA      & 0.51 & 0.33 & 0.55 \\
Moto      & 0.45 & 0.28 & 0.49 \\
MVP-LAM         & 0.58 & 0.37 & 0.60 \\
DreamDojo       & 0.55 & 0.35 & 0.58 \\
GDLAM (ours)    & \textbf{0.86} & \textbf{0.68} & \textbf{0.88} \\
\bottomrule
\end{tabular}
\end{table}

\subsection{Spatially Gated Routing Predictor}
The predictor transforms \citep{assran2023self} the current tokens $z_t$ toward the future conditioned on the latent actions $\{a_t^g\}$, as a stack of $12$ grouped DiT blocks (Fig.~\ref{fig:overview}(d)). A global condition $a_{gc}$, pooled from all group latents $[a_t^1\,a_t^2\,\cdots\,a_t^N]$, drives the shared modulation. Each block performs: (i) a global AdaLN whose six parameters (shift/scale/gate for the LN$+$SA and LN$+$MLP sub-layers) are regressed from $a_{gc}$, so shared dynamics are handled once; (ii) per-group cross-attention (LN$+$MHA), where the token stream attends to each group latent $a_t^g$ through a dedicated projection, implemented as a single batched multi-head attention over the $N$ groups so that each stream attends only to its own group; (iii) a per-group shift/scale that rescales each group's cross-attention output; and (iv) a spatial gate $\mathrm{Gate}_t$ that weights that output before it is summed back into the token stream. As shown in Fig.~\ref{fig:overview}(e), the gate pools its group latent $a_t^g$ and maps it through an MLP (Linear--GELU--Linear--GELU--Linear) to a low-resolution $14{\times}14$ mask, reshapes and applies a sigmoid, then bilinearly upsamples to the token grid, giving a per-token value in $[0,1]$. Its final layer is zero-initialized, so every gate starts closed and opens only as its group proves useful---an implicit Occam prior that lets groups specialize to distinct regions (e.g., a foreground region of change vs.\ the background). The output projection is likewise zero-initialized, so training begins at the identity map: the predictor learns a residual $\Delta$ added to $z_t$, followed by a final hyperspherical renormalization, yielding the predicted future $\hat{z}_{t+1}$.

\begin{figure}[t]
\centering
\includegraphics[width=0.95\columnwidth]{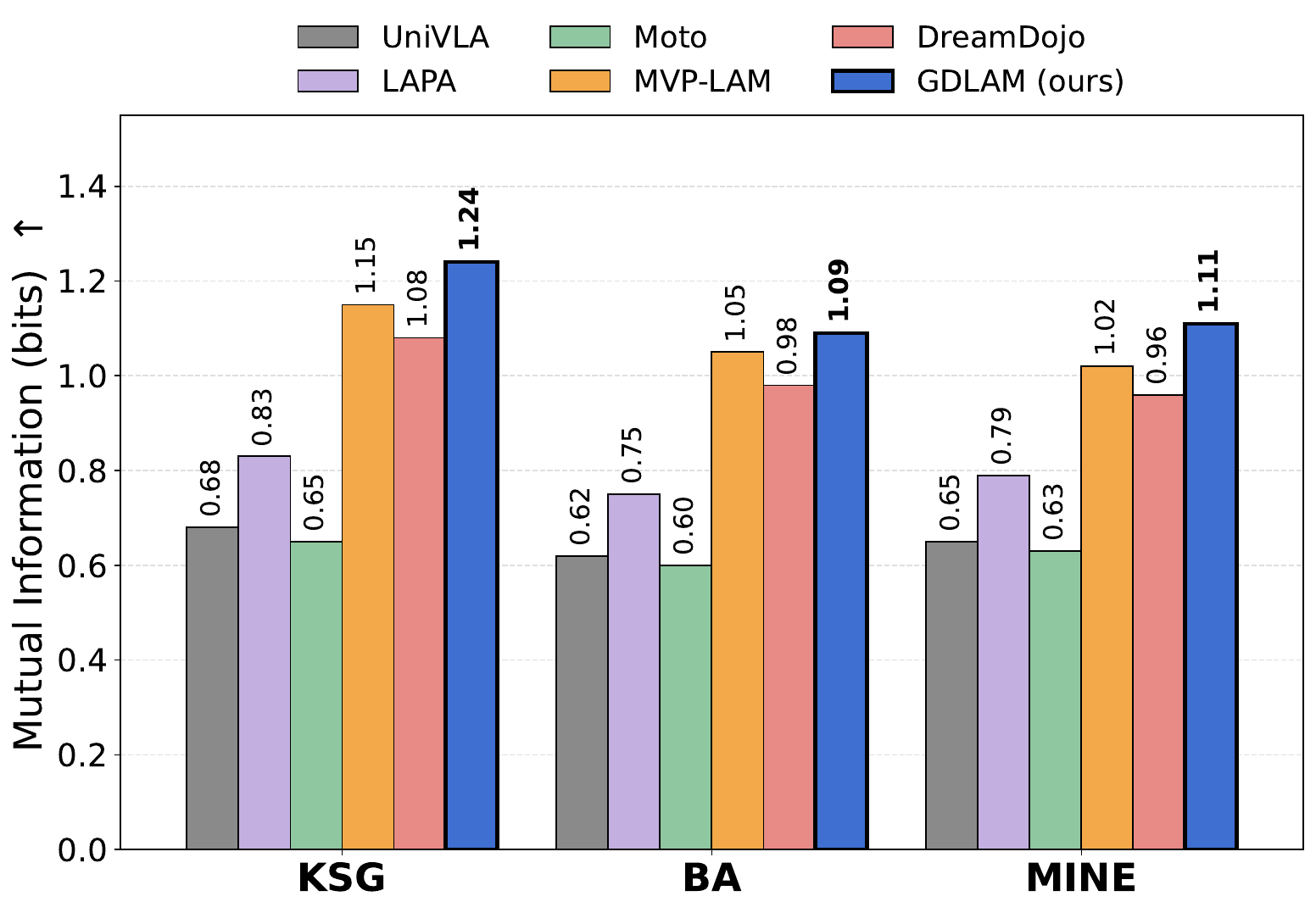}
\caption{Action-informativeness of the latent code: three MI estimators (bits, higher is better).}
\label{fig:mi}
\end{figure}

\subsection{Disentanglement Objectives}
Reconstruction alone does not force groups to capture distinct causes. On top of a reconstruction term and a per-group KL, we add four geometric objectives. Let $\mathrm{Gate}_t^{(k)}\in[0,1]^{N}$ be group $k$'s spatial gate over $G$ tokens and $a_t^{(k)}$ its latent action.

\noindent\textbf{Mutual exclusivity.} Pooling each group's latent to a summary $\bar a^{(k)}$, we penalize their pairwise cosine similarity so groups occupy near-orthogonal directions in code space:
\begin{equation}
\mathcal{L}_{\mathrm{mx}}=\frac{2}{N(N-1)}\sum_{k<k'} \Big(\tfrac{\langle \bar a^{(k)},\bar a^{(k')}\rangle}{\|\bar a^{(k)}\|\,\|\bar a^{(k')}\|}\Big)^{2}.
\end{equation}

\noindent\textbf{Group sparsity.} An $\ell_1$ penalty on the per-group latent magnitude keeps a group quiet unless its factor changes: $\mathcal{L}_{\mathrm{gs}}=\frac{1}{N}\sum_k \|a_t^{(k)}\|_1$.
 
\noindent\textbf{Gate sparsity.} An $\ell_1$ penalty on the spatial gates, $\mathcal{L}_{\mathrm{ge}}=\frac{1}{NP}\sum_k\sum_n \mathrm{Gate}_{t,n}^{(k)}$, keeps each mask compact so a group claims only the tokens it explains.

\noindent\textbf{Static--dynamic orthogonality.} We designate a single group as the static group $\mathrm{sc}$, encouraged to carry slow appearance orthogonal to change. We penalize the squared inner product between the static latent and the remaining (dynamic) latents, $\mathcal{L}_{\mathrm{orth}}=\sum_{k\neq \mathrm{sc}}\big(\langle \bar a^{(\mathrm{sc})},\bar a^{(k)}\rangle\big)^2$. This coarse time-scale split is the only imposed prior; it does not name what the dynamic groups encode.

The reconstruction term itself has three parts: a pixel loss on the decoded future frame (MSE${+}$LPIPS), an anchor loss that reconstructs the current frame from its own tokens $z_t$ (regularizing the tokenizer), and a latent loss aligning the predicted future tokens $\hat{z}_{t+1}$ with the frozen VFM's future tokens $\mathrm{sg}[z_{t+1}]$ (MSE${+}$cosine, target stop-gradiented). The total objective is
\begin{equation}
\mathcal{L}=\mathcal{L}_{\mathrm{rec}}+\beta\,\mathcal{L}_{\mathrm{KL}}+\lambda_{\mathrm{mx}}\mathcal{L}_{\mathrm{mx}}+\lambda_{\mathrm{gs}}\mathcal{L}_{\mathrm{gs}}+\lambda_{\mathrm{ge}}\mathcal{L}_{\mathrm{ge}}+\lambda_{\mathrm{orth}}\mathcal{L}_{\mathrm{orth}},
\end{equation}
with a very small $\beta$ and disentanglement weights of order $10^{-1}$--$10^{-2}$ (Appendix). To further block shortcut solutions, we apply an anti-leak augmentation: with a small probability the predictor is fed the future tokens in place of the current ones, discouraging it from copying static appearance rather than routing change through the latent action. Together these terms shape the geometry of the code: exclusivity partitions code directions, gate sparsity and exclusivity partition space, group sparsity partitions activation, and orthogonality separates time scales---jointly yielding causally distinct axes.

\begin{figure}[t]
\centering
\includegraphics[width=0.98\columnwidth]{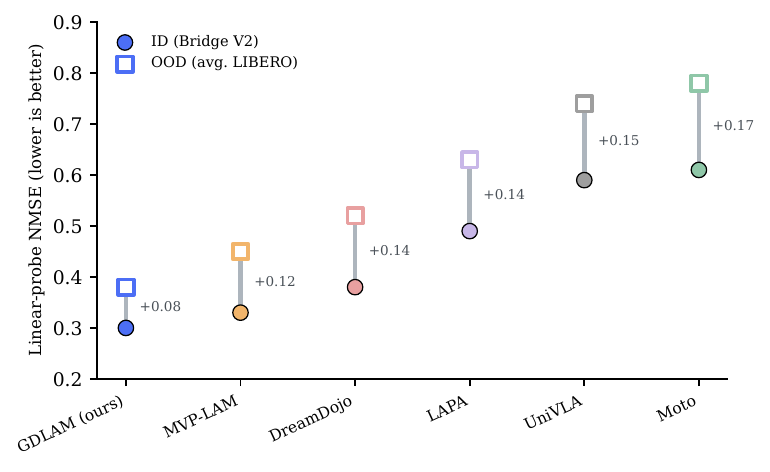}
\caption{In-distribution (ID) vs. OOD linear-probe NMSE. Each model is represented by a vertical segment connecting its ID and OOD NMSE values (square marker). Shorter segments indicate better generalization, while lower positions indicate more action-informative representations.}
\label{fig:probeood}
\end{figure}

\section{Experiments}

% 我们评估GDLAM是否学习到了一种群体解耦的潜在动作编码，以及该编码是否可作为可复用的预训练表征。我们提出三个问题：第一，GDLAM的各个群体在无监督指标下是否相互解耦，且这种解耦是否以牺牲动作信息为代价？第二，这些群体是否具有因果独立性，即干预其中一个群体是否会保持其他群体不变？第三，解耦后的编码能否在后续的世界建模与策略学习中实现连贯的迁移？关于模型选择及设计消融的补充研究详见附录。
We evaluate whether GDLAM learns a group-disentangled latent action code and whether that code behaves as a reusable pretraining representation. We ask three questions. \textbf{First}, are GDLAM's groups mutually disentangled under label-free metrics, and does this factorization cost any action information? \textbf{Second}, are the groups causally distinct, i.e.\ does intervening on one group leave the others intact? \textbf{Third}, does the disentangled code transfer coherently to downstream world modeling and policy learning? 

% 内容翻译遵循MVP-LAM的实验设置，我们在机器人数据集Open X-Embodiment以及人类操作视频数据集EgoExo4D和Assembly101上进行训练。Tokenizer采用冻结的DINOv2-with-registers ViT-Large模型；分组编码器与预测器从零开始训练。除非另有说明，$G{=}4$，$d{=}32$，编码器深度为6，预测器深度为12，训练步数约为$4.3{\times}10^{5}$（约15个周期），使用8$\times$h20 GPU。
\paragraph{Setup.} Following the experimental setup of MVP-LAM \citep{mvplam}, we train our model on the robotic dataset Open X-Embodiment \citep{o2024open} and the human manipulation video datasets EgoExo4D \citep{grauman2024ego} and Assembly101 \citep{assembly101}. The tokenizer adopts a frozen DINOv2 ViT-Large model \citep{dinov2}, while the grouped encoder and predictor are trained from scratch. Unless otherwise specified, we set the number of groups to $N{=}4$ and the latent dimension to $d{=}32$. The encoder and predictor are configured with depths of 6 and 12 layers, respectively. The model is trained for approximately $4.3{\times}10^{5}$ steps (around 15 epochs) using 8$\times$H20 GPUs. Appendix D provides more detailed parameter configurations.

% 我们与几种具有代表性的潜在动作模型（LAMs）进行对比，这些模型包括：UniVLA, LAPA, Moto, MVP-LAM, Dreamdojo（细节见附录），其中除了Dreamdojo为连续的，其他均为离散的。所有基线采用相同的数据和配置，从而确保性能差异可归因于分组码及其优化目标。
\paragraph{Baselines.} We compare GDLAM against several representative LAMs, including UniVLA \citep{univla}, LAPA \citep{lapa}, Moto \citep{moto}, MVP-LAM \citep{mvplam}, and DreamDojo \citep{gao2026dreamdojo}. Among these baselines, DreamDojo employs a continuous latent action representation, whereas the remaining methods adopt discrete latent actions. To ensure a fair comparison, all baselines are trained using the same dataset and experimental configuration. Consequently, any performance differences can be attributed to the proposed grouped latent action code and its associated optimization objectives.

\subsection{Disentanglement Without Losing Action Information}
All experiments in this section are conducted on held-out Bridge V2 transitions \citep{walke2023bridgedata}---the in-distribution setting on which every model was pretrained, disjoint from LAM pretraining---unless a shift to OOD data is stated explicitly.
% 我们采用三种标准解耦指标对每个模型进行评分，这些指标基于模型自身对编码的分区（ monolithic 基线模型为单个分区，GDLAM 为四个分区）事后计算得出：包括分组模块化得分、互信息差距（MIG）以及 DCI 得分。该评估称为“无标签”，因为模型训练未使用因子级监督，指标仅基于离线数据中的动作维度进行后验评估，用于分析潜在表示的结构化程度。完整的公式和精确的评估协议详见附录A。表 \ref{tab:disent} 显示，GDLAM 在所有基线（包括 MVP-LAM 和连续 DreamDojo 编码）上均取得显著优势：以往的 LAM 方法虽能使编码对真实动作的信息量有所增减，但并未揭示该信息在编码中如何组织，因此其解耦得分仅略高于基于 VQ 的基线模型。

\paragraph{Label-free disentanglement.} We score each model with three standard disentanglement metrics computed post-hoc against the model's own partition of the code (a single block for the monolithic baselines, four blocks for GDLAM): a per-group Modularity score, the mutual-information gap (MIG), and a DCI score. We call this evaluation label-free since no factor-level supervision is provided during training, and the metrics are computed post-hoc using offline action dimensions to assess the learned representation structure. Full formulas and the exact estimation protocol are given in Appendix~A. Table~\ref{tab:disent} shows GDLAM opens a wide margin over every baseline, including MVP-LAM and the continuous DreamDojo code: prior LAMs make the code more or less informative about the GT action, but say nothing about how that information is organized inside the code, so their disentanglement scores sit only slightly above the plain VQ-based baselines.

% 一个自然的担忧是，将编码强制划分为独立组可能会丢弃单个无约束瓶颈所保留的、与行动相关的信号。我们通过三种编码与真实网络相对行动之间的互信息估计量来检验这一点（非参数KSG估计量、Barber--Agakov变分下界以及MINE风格的Donsker--Varadhan下界），并辅以以归一化均方误差（NMSE）报告的线性探针。表~\ref{tab:info} 显示的结果与上述担忧相反：GDLAM的合并编码在所有三种估计量下都比每个基线更具信息性，且其线性探针的NMSE最低。我们将此归因于互斥性和稀疏性目标，它们阻止各组将容量浪费在相同信号的冗余副本上。
\paragraph{Does factorization cost informativeness?} A natural worry is that forcing the code into independent groups could throw away action-relevant signal that a single, unconstrained bottleneck would keep. We test this by measuring how much the frozen code reveals about the GT net-relative action, using three complementary mutual-information estimators (a non-parametric KSG estimator, a Barber--Agakov variational bound \citep{ba}, and a MINE Donsker--Varadhan bound \citep{mine}) together with a linear probe reported as normalized MSE (NMSE). Detailed definitions are provided in Appendix B. Fig.~\ref{fig:mi} shows the opposite of the worry: GDLAM's pooled code is more informative than every baseline under all three estimators, and its linear-probe NMSE is the lowest. We attribute this to the mutual-exclusivity and sparsity objectives, which discourage groups from wasting capacity on redundant copies of the same signal.

% 上述所有估计均在分布内（Bridge V2）计算得出，这也是所有模型的预训练环境。为检验GDLAM的优势是仅源于Bridge V2的特定表现，还是具备更强的泛化鲁棒性，我们在未对探针或LAM进行额外训练的条件下，于四个LIBERO套件（Spatial、Object、Goal、Long）上重复线性探针实验，即严格意义上的分布外（OOD）测试。图~\ref{fig:probeood}将各模型在分布内（Bridge V2）与分布外（LIBERO）的平均NMSE以连线形式对比呈现，线段长度直观体现了从分布内到分布外的泛化差距。GDLAM起始于最低的分布内误差，且其分布内到分布外的性能增幅最小；而朴素的单一编码基线模型初始误差较高，且在分布外进一步退化，这与内部因子化编码比单一整体编码泛化能力更强的结论一致。
\paragraph{Does the informativeness gain generalize out of distribution?} The estimates above are all in-distribution. To check whether GDLAM's advantage is a Bridge V2 artifact or a genuinely more robust signal, we move to a strictly OOD setting: four LIBERO suites \citep{libero} (Spatial, Object, Goal, Long), none seen during pretraining, on which we repeat the linear probe without any further training of the probe or the LAM. Fig.~\ref{fig:probeood} plots in-distribution against the average OOD NMSE for every model as a connected pair, so that the length of each segment directly visualizes the ID$\to$OOD generalization gap. GDLAM starts from the lowest ID error and also shows the smallest ID$\to$OOD increase, whereas the plain single-code baselines both start high and degrade further out of distribution, consistent with an internally factorized code generalizing better than a monolithic one.

\subsection{Are the Groups Causally Distinct?}
% 相关性指标可能仅满足于代码间的去相关，而非在干预下真正可分离，因此我们额外采用附录H中详述的三种协议对代码进行因果性探测：扰动单个组、从其余组预测每个组，以及系统性的成对组交换扫描。正如预告图（图\ref{fig:teaser}(a)）所示，将单组干预协议转化为跨组响应矩阵后，GDLAM得到的矩阵接近单位矩阵；我们在表\ref{tab:causal}和图\ref{fig:causal}中报告其对角线强度、平均非对角线泄漏以及跨组可预测性$R^2$。GDLAM的泄漏程度比任何基线低一个数量级。对于整体式基线而言，“组”并非明确定义的概念，因此作为参考，我们将它们的瓶颈结构一分为二，并在两个半区重复该协议；即使这些事后分割中表现最佳的结果，其矩阵的对角性也远不及GDLAM，因为其训练目标从未推动两个半区相互分离。
Correlational metrics can be satisfied by codes that are merely decorrelated rather than genuinely separable under intervention, so we additionally probe the code causally with three protocols detailed in Appendix~H: perturbing a single group, predicting each group from the remaining groups, and a systematic pairwise group-swap sweep. As previewed in the Fig.~\ref{fig:teaser}(a), condensing the single-group intervention protocol into a cross-group response matrix yields a matrix close to the identity for GDLAM; here we report its diagonal strength, mean off-diagonal leakage, and cross-group predictability $R^2$ in Table~\ref{tab:causal}. GDLAM's leakage is an order of magnitude lower than any baseline. For the monolithic baselines a ``group'' is not a defined concept, so, for reference only, we split each of their bottlenecks in half and repeat the protocol on the two halves; even the best of these post-hoc splits remains far less diagonal than GDLAM, since nothing in its training objective ever pushed the two halves apart.

\begin{figure}[t]
\centering
\includegraphics[width=0.98\columnwidth]{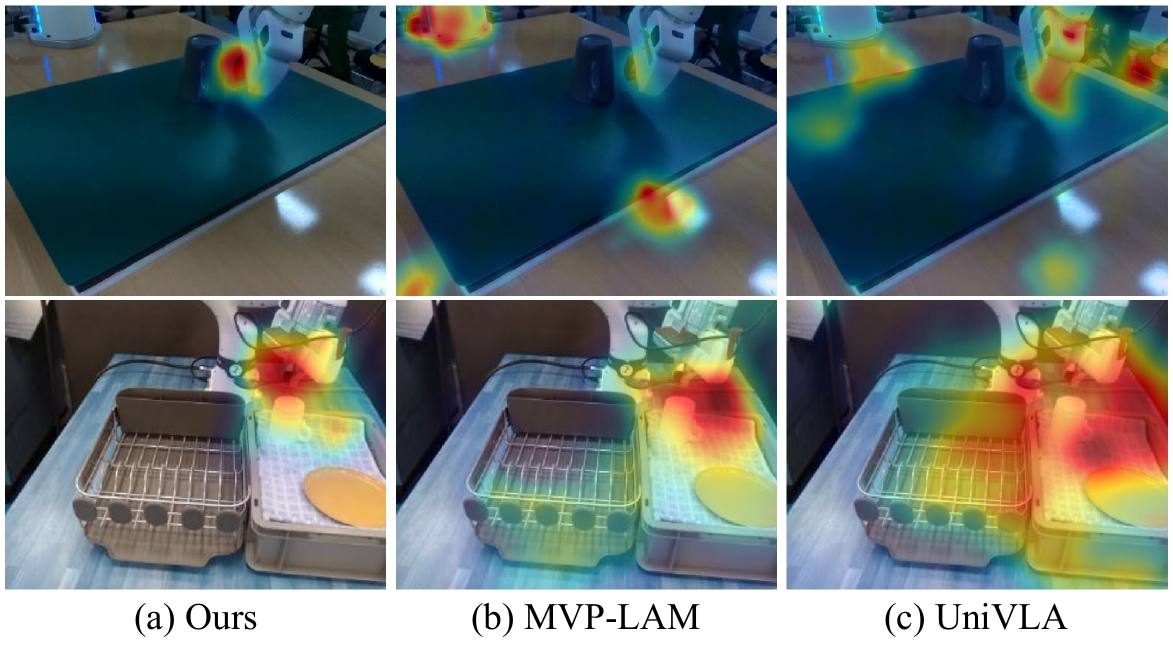}
\caption{Attention/gate localization on two representative manipulation scenes.}
\label{fig:attn}
\end{figure}

\begin{table}[t]
\centering
\caption{Causal group separation: diagonal (intended) response, off-diagonal leakage, and cross-group predictability $R^2$. For monolithic baselines the bottleneck is split in half for reference only, since these models have no trained notion of a group.}
\label{tab:causal}
\begin{tabular}{lccc}
\toprule
Model & Diag.$\uparrow$ & Leakage$\downarrow$ & Cross-$R^2\downarrow$ \\
\midrule
UniVLA    & 0.58 & 0.33 & 0.56 \\
LAPA      & 0.63 & 0.29 & 0.52 \\
Moto      & 0.56 & 0.35 & 0.58 \\
MVP-LAM         & 0.68 & 0.24 & 0.47 \\
DreamDojo       & 0.66 & 0.26 & 0.49 \\
GDLAM (ours)    & \textbf{0.91} & \textbf{0.05} & \textbf{0.14} \\
\bottomrule
\end{tabular}
\end{table}

\begin{figure}[t]
\centering
\includegraphics[width=0.8\columnwidth]{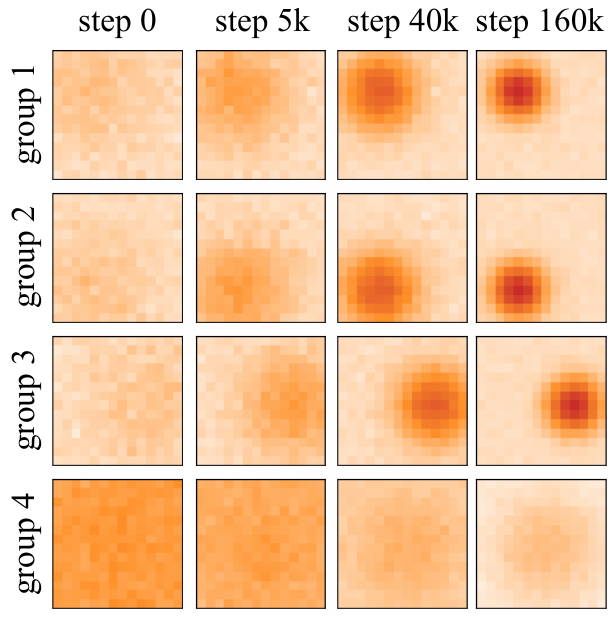}
\caption{Spatial gate maps for each of the four groups over the course of training. Columns show training progress; rows show groups. Distinct groups converge onto disjoint regions of the image without any spatial supervision.}
\label{fig:gates}
\end{figure}

\begin{table*}[htbp]
\centering
\caption{Comprehensive video evaluation results across 16 metrics spanning six dimensions. Best results are \textbf{bolded}.}
\label{tab:arena}
\resizebox{\textwidth}{!}{%
\begin{tabular}{l|ccc|ccc|ccc|cc|cc|ccc}
\toprule
\multirow{3}{*}{Methods}
  & \multicolumn{3}{c|}{\textbf{Visual Quality}}
  & \multicolumn{3}{c|}{\textbf{Motion Quality}}
  & \multicolumn{3}{c|}{\textbf{Content Consistency}}
  & \multicolumn{2}{c|}{\textbf{Physics Adherence}}
  & \multicolumn{2}{c|}{\textbf{3D Accuracy}}
  & \multicolumn{3}{c}{\textbf{Controllability}} \\
\cmidrule(lr){2-4}\cmidrule(lr){5-7}\cmidrule(lr){8-10}
\cmidrule(lr){11-12}\cmidrule(lr){13-14}\cmidrule(lr){15-17}
  & \makecell{Image\\Quality}
  & \makecell{Aesthetic\\Quality}
  & \makecell{JEPA\\Similarity}
  & \makecell{Dynamic\\Degree}
  & \makecell{Flow\\Score}
  & \makecell{Motion\\Smoothness}
  & \makecell{Subject\\Consistency}
  & \makecell{Background\\Consistency}
  & \makecell{Photometric\\Consistency}
  & \makecell{Interaction\\Quality}
  & \makecell{Trajectory\\Acc.}
  & \makecell{Depth\\Acc.}
  & Perspectivity
  & \makecell{Instruction\\Follow.}
  & \makecell{Semantic\\Align.}
  & \makecell{Action\\Follow.} \\
\midrule
GigaWorld
  & 0.463 & 0.392 & 0.437
  & 0.614 & 0.315 & 0.778
  & 0.734 & 0.832 & 0.179
  & 0.529 & 0.159
  & 0.618 & 0.760
  & 0.607 & 0.851 & 0.116 \\
Genie
  & 0.229 & 0.326 & 0.334
  & \textbf{0.674} & 0.090 & 0.694
  & 0.770 & 0.883 & 0.205
  & 0.205 & 0.074
  & 0.851 & 0.520
  & 0.208 & 0.855 & 0.018 \\
MIND-V
  & 0.327 & 0.385 & 0.460
  & 0.559 & 0.275 & 0.763
  & \textbf{0.817} & 0.909 & 0.247
  & 0.569 & 0.144
  & 0.713 & 0.796
  & 0.601 & 0.872 & 0.087 \\
RoboMaster
  & 0.353 & 0.405 & 0.302
  & 0.658 & 0.153 & 0.700
  & 0.810 & 0.903 & 0.341
  & 0.533 & 0.122
  & 0.809 & 0.738
  & 0.569 & 0.846 & 0.082 \\
Vidarc
  & 0.416 & \textbf{0.411} & 0.559
  & 0.281 & 0.145 & 0.771
  & 0.748 & 0.808 & 0.226
  & 0.528 & 0.192
  & 0.783 & 0.755
  & 0.580 & 0.863 & 0.047 \\
Cosmos 2.5
  & 0.440 & 0.355 & 0.903
  & 0.592 & 0.260 & 0.735
  & 0.804 & 0.895 & 0.355
  & 0.541 & 0.294
  & 0.862 & 0.763
  & 0.611 & 0.857 & 0.070 \\
DreamDojo (PT)
  & 0.457 & 0.376 & 0.912
  & 0.604 & 0.276 & 0.768
  & 0.806 & 0.907 & 0.360
  & 0.562 & 0.333
  & 0.873 & 0.786
  & 0.685 & 0.871 & 0.085 \\
\midrule
\textbf{Ours (PT)}
  & \textbf{0.482} & 0.394 & \textbf{0.918}
  & 0.633 & \textbf{0.319} & \textbf{0.785}
  & 0.816 & \textbf{0.914} & \textbf{0.367}
  & \textbf{0.573} & \textbf{0.350}
  & \textbf{0.884} & \textbf{0.798}
  & \textbf{0.727} & \textbf{0.894} & \textbf{0.121} \\
\bottomrule
\end{tabular}%
}
\end{table*}

% 表~\ref{tab:causal} 中的因果分离在空间门控机制中具有可见的对应表现。图~\ref{fig:gates} 追踪了各分组在训练过程中的门控状态，显示各分组逐渐收敛到互不重叠的图像区域——某些分组聚焦于紧凑的任务相关区域（如末端执行器或被操作物体），而指定的静态分组则弥散分布于背景区域——这纯粹是零初始化门控机制在降低预测误差时自动开启的结果，并未依赖任何区域标注的监督。
\paragraph{Where does the code attend?} The causal separation in Table~\ref{tab:causal} has a visible counterpart in the spatial gates. Fig.~\ref{fig:gates} tracks the gate of each group across training and shows that groups settle into disjoint image regions---some concentrate on a compact, task-relevant patch (e.g., the end effector or the object being manipulated), while the designated static group spreads diffusely over the background---purely as a consequence of the zero-initialized gates opening wherever they reduce prediction error, with no region supervision.

% 进一步，我们将GDLAM的动态组注意力与MVP-LAM及UniVLA的解码器注意力进行直接对比，三者均采用相同的注意力展开流程在相同保留的轨迹数据上计算（详见附录K）。图\ref{fig:attn}展示了两组典型场景。在两组场景中，GDLAM的注意力几乎完全集中在末端执行器上，并在接触即将发生时聚焦于即将操作的物体，且这一模式在两组场景中保持一致。MVP-LAM的定位效果虽明显优于UniVLA，但其仍会在任务无关区域形成次要热点，我们将其解释为残差、非结构化变异，单个瓶颈机制无法将其分离。
Further, we compare GDLAM's dynamic-group attention directly against the decoder attention of MVP-LAM and UniVLA, all three computed with an identical attention-rollout procedure on the same held-out transitions (Appendix~K). Fig.~\ref{fig:attn} shows two representative scenes. In both, GDLAM's attention concentrates almost entirely on the end effector and, once contact is imminent, the object it is about to manipulate, and stays that way across both scenes. MVP-LAM is noticeably better localized than UniVLA, but it still places a secondary hotspot on a task-irrelevant region of the scene, which we interpret as residual, unstructured variation that a single bottleneck has no mechanism to separate out.

\begin{table}[t]
\centering
\small
\caption{Ablations across three axes relative to the full model: disentanglement (DCI), causal separation (Leakage), and action informativeness (MI-KSG, bits).}
\label{tab:ablation}
\begin{tabular}{lccc}
\toprule
Variant & DCI$\uparrow$ & Leak.$\downarrow$ & MI$\uparrow$ \\
\midrule
Full GDLAM              & \textbf{0.88} & \textbf{0.05} & \textbf{1.22} \\
\;-- mutual excl.       & 0.61 & 0.19 & 1.14 \\
\;-- group sparsity     & 0.72 & 0.13 & 1.18 \\
\;-- gate sparsity      & 0.70 & 0.15 & 1.17 \\
\;-- static--dyn.\ orth.& 0.64 & 0.17 & 1.15 \\
shared bottleneck       & 0.57 & 0.22 & 1.10 \\
non-zero gate init      & 0.66 & 0.16 & 1.16 \\
\bottomrule
\end{tabular}
\end{table}

\subsection{Ablations}
% 表~\ref{tab:ablation} 从多个方面分析了各设计因素的贡献，所有结果均相对于完整模型进行比较。移除互斥性约束或静态--动态正交项会导致 DCI 最大幅度下降，并显著增加信息泄漏，表明这两项约束在保持组间独立性方面发挥了主要作用；组稀疏性和门控稀疏性也带来了较小但不可忽视的贡献。将多个组共享单一瓶颈，或采用非零初始化的空间门控，都会显著削弱组间专化能力，说明容量隔离机制以及基于 Occam 原则的零门控初始化并非简单的正则化技巧，而是实现结构化解耦的必要条件。值得注意的是，解耦能力和因果分离能力的提升几乎没有额外代价：动作信息量（MI）始终保持在较窄范围内。这说明所提出的结构机制主要是在重新组织潜在信息，而非简单丢弃信息。

Table~\ref{tab:ablation} analyzes the contribution of each design component, with all results compared against the full model. Removing either the mutual exclusivity constraint or the static--dynamic orthogonality objective leads to the largest degradation in DCI and the most significant increase in leakage, indicating that these two objectives play the dominant role in maintaining group separation. Group sparsity and gate sparsity also provide smaller yet non-negligible contributions. Sharing a single bottleneck across groups or initializing spatial gates away from zero substantially weakens group specialization, demonstrating that capacity isolation and the Occam-inspired zero-gate initialization are structural necessities for achieving disentanglement rather than merely auxiliary regularization techniques. Notably, the improvements in disentanglement and causal separation come with almost no additional cost: action informativeness (MI) remains within a narrow range. This suggests that the proposed structural mechanisms reorganize the information within the latent space rather than simply discarding useful information. Additional studies on choices and design ablations are reported in the Appendix I.

\subsection{Downstream Transfer}

Data scarcity remains a major bottleneck for generalization in embodied learning. Recent approaches mitigate this issue by pretraining on large-scale actionless videos and leveraging latent-action pseudo-labels extracted from them. GDLAM learns more disentangled latent action representations, which have been shown in our previous experiments to exhibit stronger alignment with GT actions. In this section, we investigate whether this structural advantage transfers to two dominant pretrain-then-posttrain paradigms in embodied intelligence: world models and VLA policies. We extract GDLAM latent-action pseudo-labels from the EgoDex dataset \citep{egodex} as the pretraining signal and evaluate the resulting models on standard downstream benchmarks. Across both settings, we keep the entire training pipeline fixed and only replace the latent action model. Full protocols, hyperparameters, and additional analyses are provided in Appendices F and G.

\paragraph{World Model.} We evaluate the GDLAM-conditioned world model on the WorldArena benchmark \citep{worldarena}, which consists of 16 metrics spanning multiple dimensions, and compare it against a broad range of representative generative models \citep{gigaworld, liao2025genie, mindv, robomaster, vidarc}. As shown in Table \ref{tab:arena}, GDLAM achieves the best overall performance, with the most significant improvements observed on the controllability dimension, including instruction following, semantic alignment, and especially action following. These results demonstrate that structured action representations effectively enhance the model’s ability to capture and utilize action-conditioned dynamics. Furthermore, we provide quantitative and qualitative evaluations on three OOD test suites in the Appendix F. GDLAM exhibits stronger action consistency and improved long-horizon prediction stability, demonstrating that disentangled latent action representations substantially improve the generalization capability of world models.

\begin{figure}[t]
    \centering
    \includegraphics[width=0.9\linewidth]{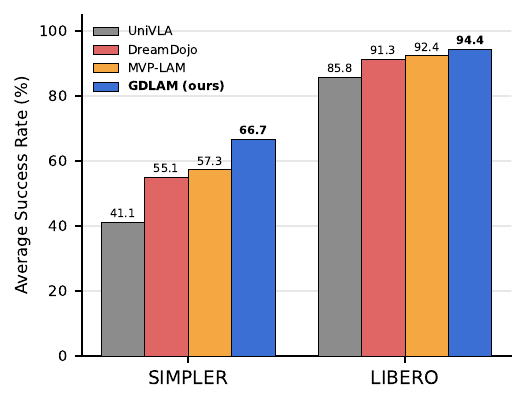}
    \caption{Average success rate of different methods across the two benchmarks.}
    \label{fig:vla}
\end{figure}

\paragraph{VLA.} Following the MVP-LAM evaluation protocol, we evaluate group-structured VLA pretraining on nine tasks across the SIMPLER \citep{simpler} and LIBERO benchmarks (Fig.~\ref{fig:vla}). GDLAM achieves the best performance on both benchmarks and all individual tasks. On SIMPLER, GDLAM reaches a $66.7\%$ average success rate, outperforming MVP-LAM by $9.4$ points. On LIBERO, GDLAM achieves a $94.4\%$ average success rate and leads across all four task families. The largest gains appear in data-limited SIMPLER and compositional LIBERO-Long ($+3.2$ over MVP-LAM), demonstrating that disentangled group latents improve control compositionality and sample efficiency (Appendix G provides detailed experiments).

% 具身领域缺乏数据，导致泛化性不足，近期利用潜在动作伪标签进行预训练提升了泛化性。GDLAM提出更具解耦的潜在动作，并通过大量实验验证与真实动作更好。那么我们通过世界模型和VLA的预训练和后训练范式进一步。我们对EgoDex数据集提取伪标签作为预训练数据，然后在大量基准上进行后训练。更多细节见附录F和G。

% 世界模型：我们在16个指标的基准WorldArena上对基于GDLAM的世界模型与大量基准，包括：GigaWorld,Genie,MIND-V,RoboMaster,Vidar,Cosmos2.5和DreamDojo进行了全面的比较。.......。此外，附录中的OOD实验和可视化分析进一步验证了我们方法在动作遵循上的优越性。显著提升泛化性。

% VLA：遵循MVP_LAM的实验设置，我们在SIMPLER和Libero基准上的9个任务上。GDLAM 在 SIMPLER 与 LIBERO 两个基准上均取得最优表现。在更具挑战性的真实机械臂 SIMPLER 基准上，GDLAM 平均成功率达到 66.7%，较次优方法 MVP-LAM（57.3%）提升 9.4 个百分点。在 LIBERO 基准上，GDLAM 平均成功率为 94.4%，在空间、物体、目标和长程四类任务上均取得最佳结果。。。。。

\section{Conclusion}
We present GDLAM, a disentangled latent action model that explicitly structures pretraining representations through grouped variational bottlenecks, gated routing, and information-geometric objectives. GDLAM learns causally distinct and robust latent action groups with improved interpretability, controllability, and transferability to world models and VLA policies.

\bibliography{aaai2027}
\clearpage
\appendix

\section{Limitations}
GDLAM still has several limitations that motivate future investigation. First, the current grouping mechanism relies on a fixed and partially hand-designed structure. Specifically, the number of latent groups $N$ and the coarse-grained static--dynamic decomposition are predefined rather than discovered from data. Consequently, the model lacks the ability to automatically adapt the granularity of factorization, such as dynamically growing, merging, or pruning latent groups according to the underlying causal factors present in different environments.

Second, the perceptual space is built upon a frozen vision foundation model. While freezing the tokenizer provides a stable and semantically meaningful representation space that facilitates training, it also inherits the intrinsic biases and perceptual limitations of the underlying foundation model. Moreover, the lack of joint optimization between visual representation learning and disentanglement objectives may restrict the model's ability to discover task-relevant factors that are not explicitly captured by the pretrained visual features.

Third, the current GDLAM framework and its downstream applications are optimized in a decoupled manner rather than through end-to-end joint training. For example, the depth--segmentation routing mechanism in the world model and the decoder mapping grouped latent actions to motor control signals in VLA policies are manually designed components rather than learned modules. Such design choices may limit the potential benefits of jointly optimizing latent representation learning, controllability, and embodied semantic alignment.

Fourth, our current evaluation primarily focuses on intrinsic representation properties and causal behavior analysis. Since GDLAM does not impose predefined semantic labels on latent groups, there is no guarantee that the discovered factors always correspond to human-interpretable concepts. Furthermore, although downstream experiments demonstrate the effectiveness of the proposed representation for world modeling and policy learning, they do not exhaustively cover the broader range of possible embodied applications. Future work will investigate more adaptive grouping strategies, jointly optimized perception--action learning frameworks, and richer evaluations of semantic interpretability and generalization.

\section{A. Label-Free Disentanglement Metrics: Definitions and Protocol}
\label{app:metrics}
Table~1 in the main text reports three quantities---Modularity \citep{M}, MIG \citep{MIG}, and DCI \citep{DCI}. All three score a code's structure against the $K{=}7$ evaluation factors of the net-relative action defined in Appendix~B (six continuous end-effector dimensions and one gripper dimension). When scoring, we treat each monolithic baseline's single code as one ``group'' spanning its full dimensionality, and for GDLAM we use the model's own four groups as the partition---without ever telling the metric which group is which.

\paragraph{Notation.} Let $z=(z_1,\dots,z_C)$ denote the $C$ scalar code units under evaluation (for GDLAM, the concatenation of all group latents; a unit is one coordinate of one group), and let $f=(f_1,\dots,f_K)$ denote the $K$ evaluation factors. We draw a held-out probing set of $(z^{(n)},f^{(n)})_{n=1}^N$ pairs from Bridge V2 \citep{walke2023bridgedata} (5{,}000 transitions, disjoint from both LAM pretraining and the MI-estimator splits in Appendix~B).

\paragraph{Modularity.} Modularity asks whether each individual code unit is informative about \emph{at most one} factor. For unit $i$ and factor $k$, we estimate the mutual information $m_{ik} = I(z_i; f_k)$ with a histogram-based discrete-continuous MI estimator (20 equal-frequency bins per continuous variable, standard for this metric). Let $k^*(i) = \arg\max_k m_{ik}$ be the factor unit $i$ is most informative about, and define the per-unit modularity score by comparing the MI profile to an idealized ``one-hot'' template that concentrates all of unit $i$'s information on $k^*(i)$:
\begin{equation}
\mathrm{Mod}_i = 1 - \frac{\sum_{k \neq k^*(i)} m_{ik}^2}{(K-1)\, m_{i,k^*(i)}^2 + \epsilon},
\end{equation}
with a small $\epsilon$ for numerical stability. The overall Modularity score is the unweighted average $\mathrm{Mod} = \frac{1}{C}\sum_i \mathrm{Mod}_i \in [0,1]$; a unit that is equally informative about every factor scores $0$, while a unit informative about exactly one factor scores $1$. For a grouped code such as GDLAM's, we additionally report the group-level version by first pooling $m_{ik}$ within each group before applying the same formula, which is the number quoted in Table~1.

\paragraph{Mutual Information Gap (MIG).} MIG asks the complementary question: for each \emph{factor}, is there a single code unit that stands out as uniquely informative, or do several units share the credit? For factor $k$, we rank code units by $m_{ik}$ and take the top two values $m_{(1)k} \geq m_{(2)k}$. The gap, normalized by the factor's own entropy $H(f_k)$ (estimated from its empirical histogram), is
\begin{equation}
\mathrm{MIG}_k = \frac{m_{(1)k} - m_{(2)k}}{H(f_k)},
\end{equation}
and the overall score is the average over factors, $\mathrm{MIG} = \frac{1}{K}\sum_k \mathrm{MIG}_k \in [0,1]$. A large gap means one unit dominates the encoding of that factor with no runner-up competing for the same information, which is the signature of disentanglement; a small gap means the factor's information is smeared across multiple units.

\paragraph{DCI (Disentanglement, Completeness, Informativeness).} DCI takes a regression-based view: we fit a gradient-boosted tree regressor $R_k(z) \approx f_k$ for each factor $k$ using the full code $z$ as input (500 trees, max depth 6, 80/20 train/test split of the probing set), and read off each tree ensemble's feature-importance vector $\{R_{ik}\}_{i=1}^C$, the relative importance of code unit $i$ for predicting factor $k$, normalized so $\sum_i R_{ik} = 1$ for each $k$. From this $C\times K$ importance matrix we compute:
\begin{align}
D_i &= 1 - H_K\!\left(\frac{R_{i1}}{\sum_k R_{ik}}, \dots, \frac{R_{iK}}{\sum_k R_{ik}}\right)\!\big/\log K, \\
D &= \sum_i \Big(\textstyle\sum_k R_{ik}\Big) D_i \Big/ \sum_{i,k} R_{ik},
\end{align}
where $H_K(\cdot)$ is the Shannon entropy of the (row-)normalized importance distribution of unit $i$ across the $K$ factors: a unit whose importance is concentrated on one factor has low entropy and high $D_i$, and the overall Disentanglement score $D$ is the importance-weighted average over units.

\begin{table}[t]
\centering
\caption{Optimization hyperparameters for the two trainable MI estimators (MINE, BA).}
\label{tab:mi-hparam}
\begin{tabular}{ll}
\toprule
Hyperparameter & Value \\
\midrule
Critic / decoder hidden width & 1024 \\
Critic / decoder depth & 4 \\
Batch size & 1024 \\
Optimizer / LR (MINE) & Adam / $1\times10^{-4}$ \\
Optimizer / LR (BA) & Adam / $5\times10^{-5}$ \\
Training steps & 8000 \\
Gradient clip & 1.0 \\
Seeds averaged & 4 \\
\bottomrule
\end{tabular}
\end{table}

\begin{figure}[t]
    \centering
    \includegraphics[width=0.98\linewidth]{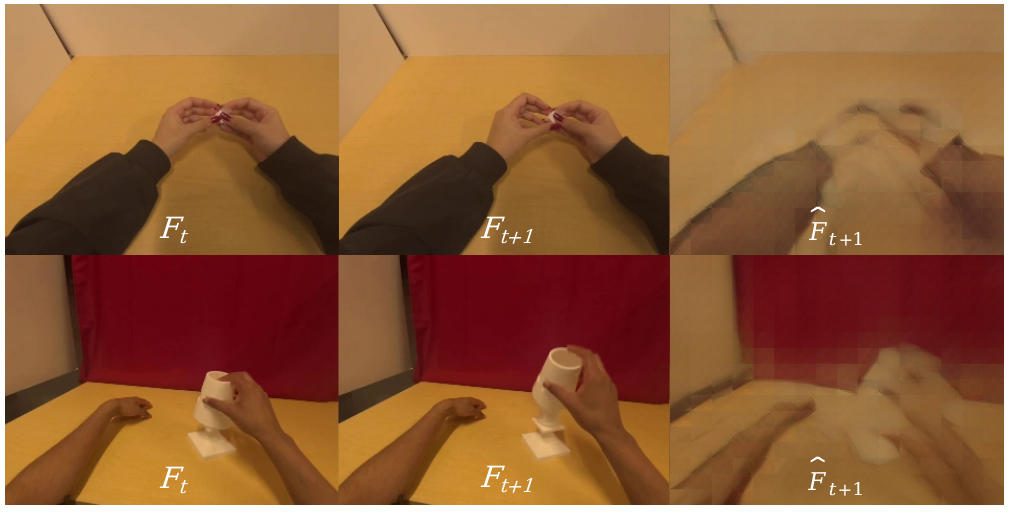}
    \caption{Pixel-space visualization of latent future prediction.}
    \label{fig:pixel}
\end{figure}

\section{B. Action Normalization and Mutual Information Protocol}
\label{app:mi-protocol}
This appendix specifies how we align GT robot actions across models, how we estimate the mutual information (MI) between latent actions and true actions, and how we run the linear probes \citep{mvplam}. Our design principle throughout is to keep the comparison \emph{capacity-invariant}: because the compared LAMs operate at different temporal strides and expose latent codes of different dimensionality, we standardize both the regression target and the estimator input so that no method gains an unfair advantage from horizon length or code width alone.

\paragraph{Horizon-consistent action target.}
Baselines predict actions over heterogeneous horizons $H$, so a raw comparison would confound the amount of motion covered by a single latent with the quality of that latent. We therefore collapse each $H$-step action chunk into one \emph{net relative action} living in a common $7$-dimensional space. Given a normalized action tensor $a^{\mathrm{norm}}\in\mathbb{R}^{B\times H\times 7}$, we first invert the dataset-level $z$-scoring per dimension,
\begin{equation}
a^{\mathrm{raw}}_{t}=a^{\mathrm{norm}}_{t}\odot\sigma+\mu,
\end{equation}
with $(\mu,\sigma)$ the per-dimension statistics of the source dataset. We then accumulate the six continuous control channels over the horizon and read off the terminal gripper state,
\begin{equation}
a^{\mathrm{net}}_{1:6}=\sum_{t=1}^{H}a^{\mathrm{raw}}_{t,1:6},\qquad
a^{\mathrm{net}}_{7}=a^{\mathrm{raw}}_{H,7}.
\end{equation}
Summation (rather than averaging) is deliberate: it preserves the cumulative displacement that the latent is supposed to explain and avoids silently rescaling the target by $1/H$. Finally, since summing $H$ approximately i.i.d.\ increments inflates the scale by $\mathcal{O}(H)$ in mean and $\mathcal{O}(\sqrt{H})$ in spread, we re-standardize with horizon-aware moments,
\begin{equation}
\hat\mu_{1:6}=H\mu_{1:6},\quad\hat\sigma_{1:6}=\sqrt{H}\,\sigma_{1:6},\quad
\hat\mu_{7}=\mu_{7},\quad\hat\sigma_{7}=\sigma_{7},
\end{equation}
\begin{equation}
a^{\mathrm{net\text{-}norm}}=(a^{\mathrm{net}}-\hat\mu)\oslash(\hat\sigma+\epsilon),
\end{equation}
where $\oslash$ is elementwise division and $\epsilon$ guards against division by zero. This yields a fixed $7$-D target whose dimensionality does not grow with $H$, so a fixed-capacity estimator sees the same problem size for every model. We adopt this identical target in both the MI estimation and the probing experiments; latent actions $z$ are held frozen throughout, so any estimator only reads out what the pretrained code already contains.

\paragraph{MI via three complementary estimators.}
Because no single MI estimator is unbiased across regimes, we triangulate $\mathcal{I}(Z;A)$ with one non-parametric and two neural lower bounds, and treat agreement in \emph{ranking} rather than absolute nats as our evidence.

\emph{(i) KSG (kNN).} We run the Kraskov--St\"ogbauer--Grassberger estimator on paired samples $\{(z^{(i)},a^{(i)})\}$ after per-dimension standardization. Since kNN density ratios degrade in high dimensions, we first apply a fixed random Gaussian projection $W\!\sim\!\mathcal{N}(0,I)$, $\tilde z=Wz\in\mathbb{R}^{256}$; because projection can only destroy information, the resulting estimate is a valid lower bound on the MI in the original code space. We use $k{=}5$ neighbors throughout.

\emph{(ii) MINE.} We fit a critic $T_\theta(z,a)$ maximizing the Donsker--Varadhan bound \citep{mine}
\begin{equation}
\mathcal{I}(Z;A)\ \ge\ \mathbb{E}_{p(z,a)}[T_\theta]-\log\mathbb{E}_{p(z)p(a)}[e^{T_\theta}],
\end{equation}
drawing product-of-marginals samples by permuting actions inside each minibatch. We average the log-partition term over several independent permutations per batch to curb its variance and report the bound on a held-out split in bits.

\emph{(iii) Barber--Agakov (BA \citep{ba}).} Writing $\mathcal{I}(Z;A)=\mathcal{H}(A)-\mathcal{H}(A|Z)$, we introduce a variational decoder $q_\phi(a|z)=\mathcal{N}(a;\mu_\phi(z),\mathrm{diag}(\sigma^2))$ with an MLP mean and a learned global scale, giving
\begin{equation}
\mathcal{I}(Z;A)\ \ge\ \mathcal{H}(A)+\mathbb{E}_{p(z,a)}[\log q_\phi(a|z)].
\end{equation}
We fit $\phi$ by maximum likelihood, approximate the marginal entropy with a Gaussian $q(a)$ fitted to the training actions, and report the plug-in estimate $\widehat{\mathcal{I}}_{\mathrm{BA}}=\frac{1}{\log 2}\big(\mathbb{E}_{p(z,a)}[\log q_\phi(a|z)]-\mathbb{E}_{p(a)}[\log q(a)]\big)$ in bits.

\paragraph{Estimator protocol and sanity checks.}
The two neural estimators (MINE, BA) are trained on a training split with early stopping selected on a disjoint validation split, and evaluated once on a held-out test split; every number is averaged over four seeds controlling both data resampling and optimization noise, and we report mean$\pm$std. Table~\ref{tab:mi-hparam} lists the concrete optimization hyperparameters for both critics. As a negative control, we shuffle the $(z,a)$ pairing at test time and verify that all three estimators collapse toward zero dependence, confirming they measure genuine coupling rather than fitting artifacts. To rule out the trivial explanation that MI gaps merely reflect different code diversity, we additionally report the empirical Shannon entropy $\hat{\mathcal{H}}(Z)$ of the quantized codes on the same subset.

\paragraph{Linear probing.}
For each dataset we build a probing set $\{(z^{(i)},a^{(i)})\}$ and fit a \emph{single} affine layer $\hat a=Wz+b$ by minimizing $\mathbb{E}\|a-\hat a\|_2^2$, with the LAM frozen. To keep probe capacity constant across codes of differing width, we first reduce each latent to $d{=}128$ PCA components (for GDLAM we concatenate all group codes before PCA so the grouped code is not penalized or advantaged by its factorization). We report the normalized MSE, $\mathrm{NMSE}=\mathbb{E}\|a-\hat a\|_2^2/\mathrm{Var}(a)$, so that a value of $1$ corresponds to predicting the mean action. A low probe NMSE indicates the code linearly exposes step-level control signals, complementing the MI estimates which also capture nonlinear dependence. For GDLAM specifically, all three MI estimators and the linear probe are applied to the \emph{pooled} code (all groups concatenated), never to a single group in isolation, so that the resulting numbers answer whether factorization costs any of the total action information available to a monolithic code; per-group informativeness is instead addressed by the causal intervention protocols in Appendix~H.

\begin{figure*}[t]
    \centering
    \includegraphics[width=1\linewidth]{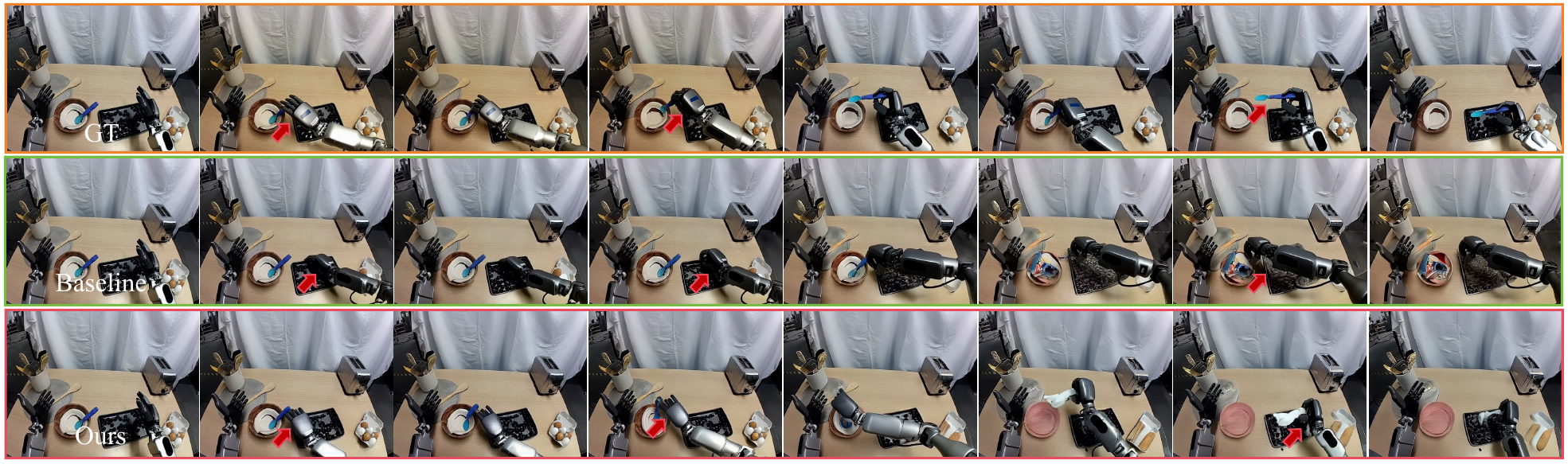}
    \caption{Qualitative visualization of world model rollouts on the DreamDojo-HV test set (30-frame intervals).}
    \label{fig:roll}
\end{figure*}

\begin{table}[t]
\centering
\caption{Next-frame prediction error in the shared, frozen DINOv2 feature space (pixel-decoding baselines are re-embedded post-hoc). Lower is better for MSE, higher is better for cosine similarity.}
\label{tab:dinofid}
\begin{tabular}{lcc}
\toprule
Model & DINOv2 MSE$\downarrow$ & Cosine sim.$\uparrow$ \\
\midrule
UniVLA          & 0.148 & 0.881 \\
LAPA            & 0.162 & 0.865 \\
Moto            & 0.171 & 0.851 \\
MVP-LAM               & 0.129 & 0.902 \\
DreamDojo             & 0.133 & 0.898 \\
GDLAM (ours)          & \textbf{0.126} & \textbf{0.905} \\
\bottomrule
\end{tabular}
\end{table}

\begin{table}[t]
\centering
\caption{Main hyperparameters. Loss weights are annealed; see the released config for schedules.}
\label{tab:hparam}
\begin{tabular}{ll}
\toprule
Hyperparameter & Value \\
\midrule
Number of groups $N$ & 4 (swept: 1, 2, 4, 8) \\
Tokens per group $\tau$ & 1 \\
Latent dim $d$ & 32 \\
Encoder depth & 6 \\
Predictor depth & 12 \\
Tokenizer & DINOv2-reg ViT-L (frozen) \\
KL weight $\beta$ & $10^{-4}\!\to\!10^{-3}$ (annealed) \\
Mutual-exclusivity $\lambda_{\mathrm{mx}}$ & $0\!\to\!0.1$ (annealed) \\
Group-sparsity $\lambda_{\mathrm{gs}}$ & $0\!\to\!0.05$ (annealed) \\
Gate-sparsity $\lambda_{\mathrm{ge}}$ & $0\!\to\!0.05$ (annealed) \\
Orthogonality $\lambda_{\mathrm{orth}}$ & $0\!\to\!0.1$ (annealed) \\
Batch size & 64 \\
Optimizer / LR & AdamW / $2\times10^{-4}$, cosine \\
Weight decay & $10^{-2}$ \\
Gradient clip & 1.0 \\
Training steps & $\approx 4.3{\times}10^{5}$ \\
Hardware & 8$\times$h20 (96GB) \\
Wall-clock & $\approx$ 6.5 days \\
\bottomrule
\end{tabular}
\end{table}

\section{C. Fidelity in a Shared Perceptual Space}
\label{sec:dinofid}
% 在我们比较的基线模型中，并非所有模型都解码到相同的输出空间：Moto风格模型直接解码RGB像素，LAPA风格和UniVLA风格模型则对像素邻域特征网格进行量化与重建，而MVP-LAM和GDLAM均预测下一帧的冻结DINOv2特征向量，完全不涉及像素层面。因此，PSNR等像素空间指标实际上是在比较不同范畴的结果：像素解码器因能生成精细纹理而获得奖励，而特征空间预测器从未被要求生成此类信息，且这种奖励与两种方法在下游任务中的实际效用无关。为此，我们采用统一评估标准：对于像素解码基线，我们使用相同的冻结DINOv2编码器对其预测帧进行重新编码；对所有模型，我们报告预测帧与真实下一帧的DINOv2特征向量之间的差异，以均方误差和余弦相似度衡量。表\ref{tab:dinofid}显示，GDLAM在此次比较中取得了最佳保真度，与MVP-LAM和DreamDojo表现相当或略胜一筹；而像素解码基线（尤其是Moto风格模型）在特征空间误差上明显更大，这与其解码器将容量用于纹理而非下游策略或世界模型实际依赖的语义内容相一致。
The baselines in our comparison do not all decode into the same output space: Moto decodes RGB pixels directly, LAPA and UniVLA quantize and reconstruct a pixel-adjacent feature grid, while MVP-LAM and GDLAM both predict the next frame's frozen DINOv2 tokens. Pixel-space metrics such as PSNR therefore compare apples to oranges: a pixel decoder is rewarded for fine-grained texture that a feature-space predictor was never asked to produce, independent of how useful either code is downstream. We instead standardize every model onto one yardstick: for pixel-decoding baselines we re-embed their predicted frame with the same frozen DINOv2 encoder used everywhere else, and for every model we report the discrepancy between the predicted and GT DINOv2 tokens of the true next frame, as mean-squared error and cosine similarity. Table~\ref{tab:dinofid} shows GDLAM gives the best fidelity in the comparison, matching or edging out MVP-LAM and DreamDojo; the pixel-decoding baselines (Moto in particular) incur a visibly larger feature-space error, consistent with their decoders spending capacity on texture rather than on the semantic content a downstream policy or world model actually consumes.

Although GDLAM incorporates a pixel-level reconstruction constraint and an auxiliary pixel decoder, we emphasize that pixel-perfect reconstruction is not the objective of our method. Since the latent tokens are derived from low-dimensional DINOv2 features, they inherently discard fine-grained appearance details and are not designed to support faithful pixel-level image recovery. As illustrated in Fig.~\ref{fig:pixel}, the pixel-space visualizations obtained through the auxiliary decoder nevertheless reveal that GDLAM produces more accurate and coherent motion predictions, while further suppressing background-related and appearance-irrelevant pixel information. This observation suggests that the learned latent actions capture the underlying dynamic structure governing future changes, rather than merely performing appearance-driven regression toward averaged visual outcomes.

\section{D. Architecture and Hyperparameters}
Table~\ref{tab:hparam} lists the main hyperparameters; below we walk through them and explain the reasoning behind the key choices.

\paragraph{Tokenizer and encoder--predictor asymmetry.} The visual backbone is a frozen DINOv2-with-registers ViT-L: freezing it forces the latent action to encode \emph{change} between frames rather than re-learn appearance, and its patch tokens are $\ell_2$-normalized onto the hypersphere so that no high-norm token dominates the cross-attention delta computation. The encoder ($6$ blocks) is deliberately much shallower than the predictor ($12$ blocks): the encoder only has to compress the inter-frame difference into the compact per-group code, whereas the predictor bears the heavier task of reconstructing future dynamics from that code, so we allocate roughly twice the depth to it.

\paragraph{Grouped bottleneck.} The code is factorized into $N{=}4$ groups (one slow static/scene group and three dynamic groups), a value we did not pick arbitrarily but selected from the sweep $N\in\{1,2,4,8\}$ (Appendix~I); $N{=}4$ balances disentanglement against capacity. Each group gets $\tau{=}1$ token of latent dimension $d{=}32$, an intentionally tight bottleneck: a single low-dimensional slot per group discourages a group from absorbing multiple factors and gives the mutual-exclusivity and sparsity penalties a well-defined target. A learnable per-group prefix embedding keeps the groups distinguishable at the input while they share the encoder trunk. Numerically, the variational reparameterization runs in fp32 to avoid the variance collapse that low-precision sampling induces in the KL term, and the spatial gates and predictor output projection are zero-initialized so training starts from an identity map.

\paragraph{Loss weights and annealing.} All four disentanglement terms and the KL weight are annealed from (near) zero to their targets rather than applied at full strength: the KL weight $\beta$ ramps $10^{-4}\!\to\!10^{-3}$, mutual-exclusivity $\lambda_{\mathrm{mx}}$ and orthogonality $\lambda_{\mathrm{orth}}$ each rise to $0.1$, and the two sparsity terms $\lambda_{\mathrm{gs}},\lambda_{\mathrm{ge}}$ to $0.05$. The ordering of magnitudes is deliberate---the structural terms that shape \emph{which} group owns a factor ($\lambda_{\mathrm{mx}},\lambda_{\mathrm{orth}}$) are weighted twice as strongly as the sparsity regularizers that merely keep each group quiet---and starting them all near zero lets reconstruction establish itself before the penalties compete with it, which otherwise yields degenerate, information-poor codes. Exact schedules are in the released config.

\paragraph{Optimization and compute.} We train with AdamW at a peak learning rate of $2\times10^{-4}$ under a cosine schedule, weight decay $10^{-2}$, gradient clipping at $1.0$, and batch size $64$, for $\approx4.3\times10^{5}$ steps. The full run takes $\approx6.5$ days on $8\times$H20 (96\,GB) GPUs; this cost, driven mainly by the predictor depth and the frozen high-resolution tokenizer, is the practical reason we treat larger $N$ and larger-scale data as future work rather than exhaustively re-sweeping every setting.

\begin{figure*}
    \centering
    \includegraphics[width=1\linewidth]{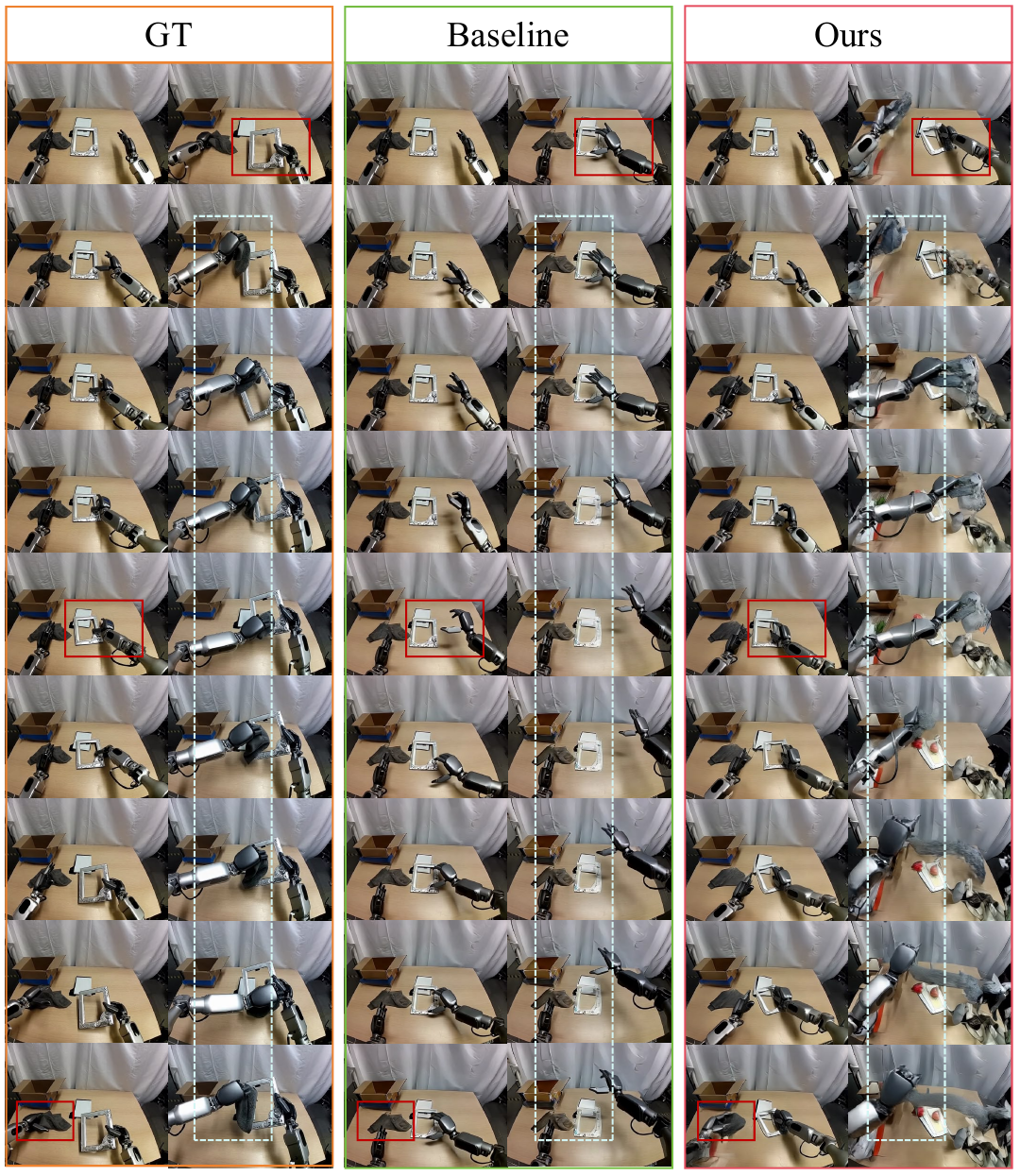}
    \caption{Long-horizon rollout visualization results of the world model on the DreamDojo-HV test set (with frames sampled every 20 steps).}
    \label{fig:long}
\end{figure*}

\begin{table*}[t]
\centering
\caption{World-model OOD transfer with the DreamDojo generator held fixed, varying only the latent action model that supplies the conditioning code.}
\label{tab:wm}
\begin{tabular}{lccccccccc}
\toprule
& \multicolumn{3}{c}{In-lab} & \multicolumn{3}{c}{EgoDex} & \multicolumn{3}{c}{DreamDojo-HV} \\
\cmidrule(lr){2-4}\cmidrule(lr){5-7}\cmidrule(lr){8-10}
Conditioning code & PSNR$\uparrow$ & SSIM$\uparrow$ & LPIPS$\downarrow$ & PSNR$\uparrow$ & SSIM$\uparrow$ & LPIPS$\downarrow$ & PSNR$\uparrow$ & SSIM$\uparrow$ & LPIPS$\downarrow$ \\
\midrule
DreamDojo (retrained) & 20.42 & 0.761 & 0.231 & 19.78 & 0.770 & 0.229 & 18.19 & 0.740 & 0.246 \\
MVP-LAM               & 20.57 & 0.766 & 0.227 & 19.91 & 0.775 & 0.225 & 18.31 & 0.745 & 0.242 \\
GDLAM (ours)          & \textbf{20.88} & \textbf{0.774} & \textbf{0.221} & \textbf{20.14} & \textbf{0.782} & \textbf{0.219} & \textbf{18.53} & \textbf{0.751} & \textbf{0.236} \\
\bottomrule
\end{tabular}
\end{table*}

\section{E. Disentanglement Objectives: Details}
Let $\mathrm{Gate}_t^{(k)}\in[0,1]^{P}$ be the spatial gate of group $k$ over $P$ tokens and $a_t^{(k)}\in\mathbb{R}^{d}$ its latent action, with $\bar a^{(k)}$ its pooled summary. $\mathcal{L}_{\mathrm{rec}}$ combines a pixel loss on the decoded future frame, an anchor loss on the current frame, and a cosine/MSE latent loss between predicted future tokens $\hat{z}_{t+1}$ and $\mathrm{sg}[z_{t+1}]$; $\mathcal{L}_{\mathrm{KL}}$ is the per-group Gaussian KL to a unit prior. The four disentanglement terms are:
\begin{align}
\mathcal{L}_{\mathrm{mx}} &= \tfrac{2}{N(N-1)}\textstyle\sum_{k<k'}\big(\cos(\bar a^{(k)},\bar a^{(k')})\big)^2, \\
\mathcal{L}_{\mathrm{gs}} &= \tfrac{1}{N}\textstyle\sum_{k}\|a_t^{(k)}\|_1, \\
\mathcal{L}_{\mathrm{ge}} &= \tfrac{1}{NP}\textstyle\sum_{k}\sum_{n} \mathrm{Gate}_{t,n}^{(k)}, \\
\mathcal{L}_{\mathrm{orth}} &= \textstyle\sum_{k\neq \mathrm{sc}}\big(\langle \bar a^{(\mathrm{sc})},\bar a^{(k)}\rangle\big)^2,
\end{align}
where $\cos(\cdot,\cdot)$ is cosine similarity. Intuitively, $\mathcal{L}_{\mathrm{mx}}$ pushes group latents toward orthogonal directions; $\mathcal{L}_{\mathrm{gs}}$ keeps each group quiet unless its factor changes; $\mathcal{L}_{\mathrm{ge}}$ keeps each spatial gate compact; $\mathcal{L}_{\mathrm{orth}}$ keeps the slow \emph{scene} factor orthogonal to motion. When the cross-viewpoint auxiliary is enabled, $\mathcal{L}_{\mathrm{xv}}$ is added to $\mathcal{L}_{\mathrm{rec}}$ with the same weight as the self-view term, and no additional disentanglement term is introduced, since the auxiliary acts purely on the pooled global condition $a_{gc}$.

\begin{table}[t]
\centering
\caption{VLA pretraining and finetuning hyperparameters.}
\label{tab:vla-hparam}
\begin{tabular}{ll}
\toprule
Hyperparameter & Value \\
\midrule
VLM backbone & Prismatic-7B \\
Pretrain steps & 200k \\
Pretrain LR & $2\times10^{-5}$ \\
Pretrain batch size & 96 \\
SIMPLER FT & 10k steps, bs=4, accum=4 \\
LIBERO FT & 30k steps, bs=8, accum=2 \\
LoRA $(r,\alpha)$ & (32,16) \\
Action horizon $H$ & 5 / 12 \\
\bottomrule
\end{tabular}
\end{table}

\begin{table*}[t]
\centering
\caption{SIMPLER benchmark. Success rate (\%) on a 7-DoF WidowX arm.}
\label{tab:simpler}
\begin{tabular}{lccccc}
\toprule
Method & StackG2Y & Carrot2Plate & Spoon2Towel & Eggplant2Bask & AVG \\
\midrule
UniVLA VLA & 18.7 & 33.3 & 54.2 & 58.3 & 41.1 \\
DreamDojo VLA     & 27.1 & 58.3 & 66.7 & 68.3 & 55.1 \\
MVP-LAM VLA       & 29.2 & 62.5 & 66.7 & 70.8 & 57.3 \\
GDLAM VLA (ours)  & \textbf{37.5} & \textbf{75.0} & \textbf{75.0} & \textbf{79.2} & \textbf{66.7} \\
\bottomrule
\end{tabular}
\end{table*}

\begin{table*}[t]
\centering
\caption{LIBERO benchmark. Success rate (\%) for VLAs pretrained on the same Bridge V2 + OXE mixture.}
\label{tab:libero}
\begin{tabular}{lccccc}
\toprule
Method & Spatial & Object & Goal & Long & AVG \\
\midrule
UniVLA VLA & 89.7 & 90.2 & 85.1 & 78.3 & 85.8 \\
DreamDojo VLA     & 94.6 & 94.1 & 90.8 & 85.7 & 91.3 \\
MVP-LAM VLA       & 95.4 & 94.8 & 91.6 & 87.9 & 92.4 \\
GDLAM VLA (ours)  & \textbf{96.8} & \textbf{96.2} & \textbf{93.5} & \textbf{91.1} & \textbf{94.4} \\
\bottomrule
\end{tabular}
\end{table*}

\section{F. World Model: Details}
\label{app:wm}
To show that GDLAM's disentangled code transfers as a pretraining signal for world modeling, we do not design a new world-model architecture. Instead, we take the DreamDojo world model unchanged and only replace its latent action model with GDLAM, so that any difference is attributable to the conditioning code rather than to the generator.

\paragraph{Backbone and conditioning.} Following DreamDojo, the world model is a latent video diffusion transformer (Cosmos-Predict2.5) operating in the continuous latent space of a WAN2.2 tokenizer and trained with a flow-matching objective augmented by DreamDojo's temporal-consistency loss ($\lambda{=}0.1$). The latent action is injected exactly as in DreamDojo: it is projected by a lightweight MLP (whose last layer is zero-initialized) to the dimension of the timestep embedding, added to that embedding, and consumed by the adaptive layer normalization (AdaLN) of every DiT block \citep{dit}. We keep DreamDojo's two conditioning designs untouched---relative actions rebaselined per latent frame, and chunked action injection where the four actions spanning a latent frame are sent to that frame together. The only change is the source of the latent action: DreamDojo's own monolithic continuous code is swapped for GDLAM's code.

\paragraph{Injecting the grouped code.} Because GDLAM emits $N$ group latents $\{a_t^{(g)}\}$ rather than a single vector, we pool them into one conditioning vector $a_t=\mathrm{concat}(a_t^{(1)},\dots,a_t^{(N)})$ before the DreamDojo action MLP, so the injection interface, its dimensionality, and the zero-initialized last layer are all identical to the baseline.

\paragraph{Pretraining and post-training.} We follow DreamDojo's pretrain--post-train recipe and change only the latent action source. For pretraining, we re-train DreamDojo's LAM stage---i.e.\ we substitute GDLAM for its original latent action model---on the same data used throughout this paper (the Setup in the main text), and condition the world model on the resulting latent actions via the interface above. We then post-train the world model on GR1\_robot dataset and evaluate zero-shot on three held-out OOD suites---In-lab, EgoDex, and DreamDojo-HV---whose objects, scenes, and interactions are unseen during GR1\_robot post-training \citep{gao2026dreamdojo}.

\paragraph{Comparison.} We compare world models that share this identical DreamDojo pipeline and differ only in the LAM that provides the conditioning code: DreamDojo's own monolithic continuous code, MVP-LAM, and GDLAM (ours). As shown in Table~\ref{tab:wm}, conditioning the DreamDojo world model on GDLAM's disentangled code improves OOD rollout fidelity on every suite and metric. Averaged over the three suites, GDLAM improves PSNR by $+0.39$ over DreamDojo and $+0.25$ over MVP-LAM while lowering LPIPS by $0.010$ and $0.006$, respectively.

Furthermore, we provide qualitative OOD comparisons on the DreamDojo-HV dataset, as shown in Fig.~\ref{fig:roll} and Fig.~\ref{fig:long}. Compared with the baseline methods, our approach achieves superior action adherence and higher generation fidelity. As illustrated in Fig.~\ref{fig:roll}, the rollout conditioned on GDLAM produces more distinct and well-structured finger configurations for the right hand, while maintaining a motion trajectory that is more consistent with the GT future states. This demonstrates that the disentangled latent action representation provides more precise motion-related guidance, enabling the world model to better capture fine-grained manipulation dynamics. Fig.~\ref{fig:long} further presents longer-horizon rollout visualizations. In the bimanual manipulation scenarios, baseline models struggle to preserve the continuity of the left-hand motion, often resulting in missing or nearly static left-hand actions. In contrast, GDLAM maintains more prominent and temporally consistent movements for both hands, producing trajectories that better match the actual evolution of the scene. These results suggest that GDLAM's structured latent action representation effectively captures independent variations among different motion factors, providing a more reliable conditioning signal for long-horizon controllable generation.

We note that the background distortions observed in the rollouts mainly originate from the limitations of the generative world model itself rather than the latent action model. Such artifacts typically arise under long-range occlusions and complex scene transitions, reflecting the inherent challenges of visual generation and memory modeling. In comparison, the primary advantage of GDLAM lies in its improved modeling of action-related dynamics: by learning a more disentangled and semantically structured action representation, GDLAM provides cleaner motion conditions that enable the unchanged generator to better follow the target actions.

\begin{figure*}[t]
\centering
\includegraphics[width=0.95\linewidth]{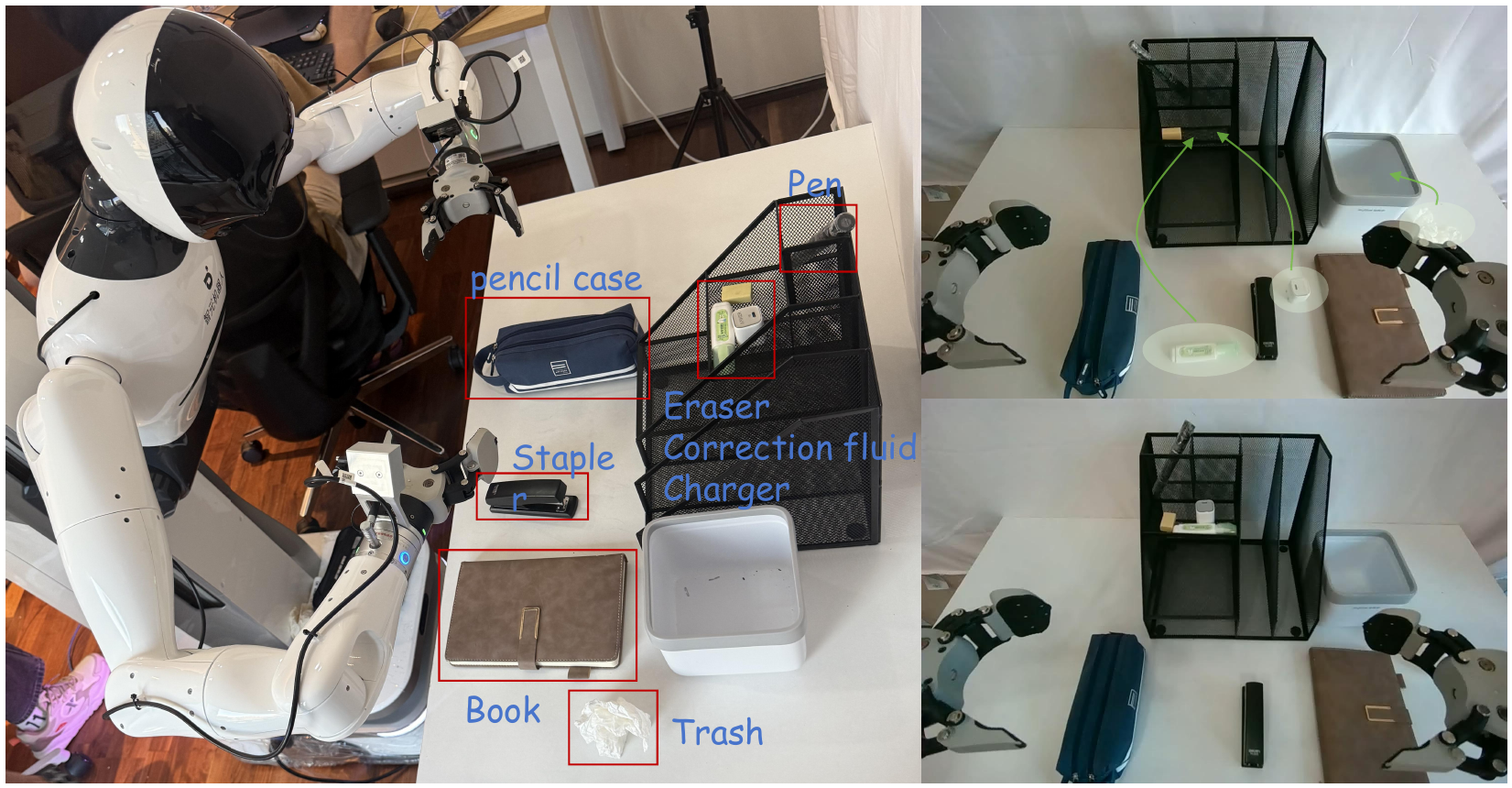}
\caption{Real-robot deployment on an AgiBot G1 in the desk scene.}
\label{fig:realbot}
\end{figure*}

\begin{table*}[t]
\centering
\small
\caption{Language instructions for the eight desktop subtasks in \textbf{g1\_all\_realbot}. \emph{Arm} denotes the acting arm (L/R).}
\label{tab:realbot-tasks}
\begin{tabular}{llp{12.4cm}}
\toprule
Split & Arm & Instruction \\
\midrule
\texttt{g1\_pen}        & L & The left arm picks up the pen on the desk, and the left arm places the pen into the pen holder at the top of the storage box. \\
\texttt{g1\_eraser}     & R & The right arm picks up the eraser on the desk, and the right arm places the eraser on the third layer of the storage box from the bottom. \\
\texttt{g1\_correction} & L & The left arm picks up the correction fluid on the desk, and the left arm places the correction fluid on the third layer of the storage box from the bottom. \\
\texttt{g1\_charger}    & R & The right arm picks up the charger on the desk, and the right arm places the charger on the third layer of the storage box from the bottom. \\
\texttt{g1\_pencil}     & L & The left arm picks up the pencil case on the desk, and the left arm places the pencil case on the second layer of the storage box from the bottom. \\
\texttt{g1\_stapler}    & R & The right arm picks up the stapler on the desk, and the right arm places the stapler on the second layer of the storage box from the bottom. \\
\texttt{g1\_books}      & R & The right arm pushes the books on the desk to the edge of the desk, and the right arm picks up the books on the desk, and the right arm places the books on the bookshelf beside the storage box. \\
\texttt{g1\_trash}      & R & The right arm picks up the trash on the desk, and the right arm places the trash into the trash can. \\
\bottomrule
\end{tabular}
\end{table*}

\section{G. Group-Structured VLA Pretraining}
\label{app:vla}
To show that GDLAM's disentangled code transfers as a pretraining signal for policy learning, we follow the UniVLA latent action VLA pipeline unchanged and only replace the latent action model with GDLAM.

\paragraph{Backbone and pretraining.} Following MVP-LAM, the backbone is a Prismatic-7B VLM \citep{kim2024openvla} that ingests observation tokens and a language instruction. We adopt their latent action pretraining recipe unchanged and pretrain it on actionless video paired with instructions to predict GDLAM's stop-gradient latent action targets, one prediction head per group, under $\mathcal{L}_{\mathrm{vla}}=\sum_g w_g\|\hat a_t^{(g)}-\mathrm{sg}[a_t^{(g)}]\|_2^2$. The monolithic baselines (UniVLA, DreamDojo, MVP-LAM) are pretrained identically except that the VLM emits a single vector matched to their monolithic code; backbone, optimizer, and pretraining budget are held fixed across all variants (Table~\ref{tab:vla-hparam}).

\paragraph{Injecting the grouped code.} Because GDLAM emits $N$ group latents rather than a single vector, we attach $N$ group heads $\{\hat a_t^{(g)}\}$ and fit a group-to-motor decoder $D:\{a_t^{(g)}\}\!\to\!\mathbf{a}$ end-to-end, so each group is routed to the degrees of freedom it drives without pre-assigning a fixed semantic role. Following UniVLA's multi-head attention action-prediction design, the monolithic baselines use the same-capacity decoder to map their single vector to $\mathbf{a}$, keeping the control interface and its capacity identical across variants.

\paragraph{Finetuning and evaluation.} We convert the pretrained VLM into a VLA with LoRA and, following the evaluation protocol of MVP-LAM, test on two standard manipulation benchmarks. SIMPLER evaluates four tasks on a 7-DoF WidowX arm; since it provides no official finetuning set, we follow MVP-LAM and finetune on the same 100 demonstrations (25 per task) used in their setup, reporting success rate. LIBERO comprises four suites (Spatial, Object, Goal, Long), each with 10 tasks, on which we report the average success rate under the same protocol. All VLAs share the same Bridge V2 + OXE pretraining mixture and the same finetuning steps (Table~\ref{tab:vla-hparam}), so the comparison isolates the effect of the latent action model.

\paragraph{Comparison.} Tables~\ref{tab:simpler}--\ref{tab:libero} report per-task results, and across both suites the group-structured VLA is the best method on every task. The gap over UniVLA is large throughout ($+25.6$ on SIMPLER and $+8.6$ on LIBERO in average success rate), consistent with its comparatively weak action-informativeness. The more informative comparison is against the strongest baseline, MVP-LAM: GDLAM raises the SIMPLER average from $57.3$ to $66.7$ ($+9.4$) and the LIBERO average from $92.4$ to $94.4$ ($+2.0$).

A factorized code helps most when control must be composed or is data-starved, GDLAM's advantage is largest on SIMPLER (finetuned with only 100 demonstrations) and on LIBERO-Long ($+3.2$ over MVP-LAM), which most requires composing sub-behaviors in sequence, whereas the near-saturated LIBERO-Spatial and LIBERO-Object suites leave little headroom.

\begin{figure*}[t]
    \centering
    \includegraphics[width=0.98\linewidth]{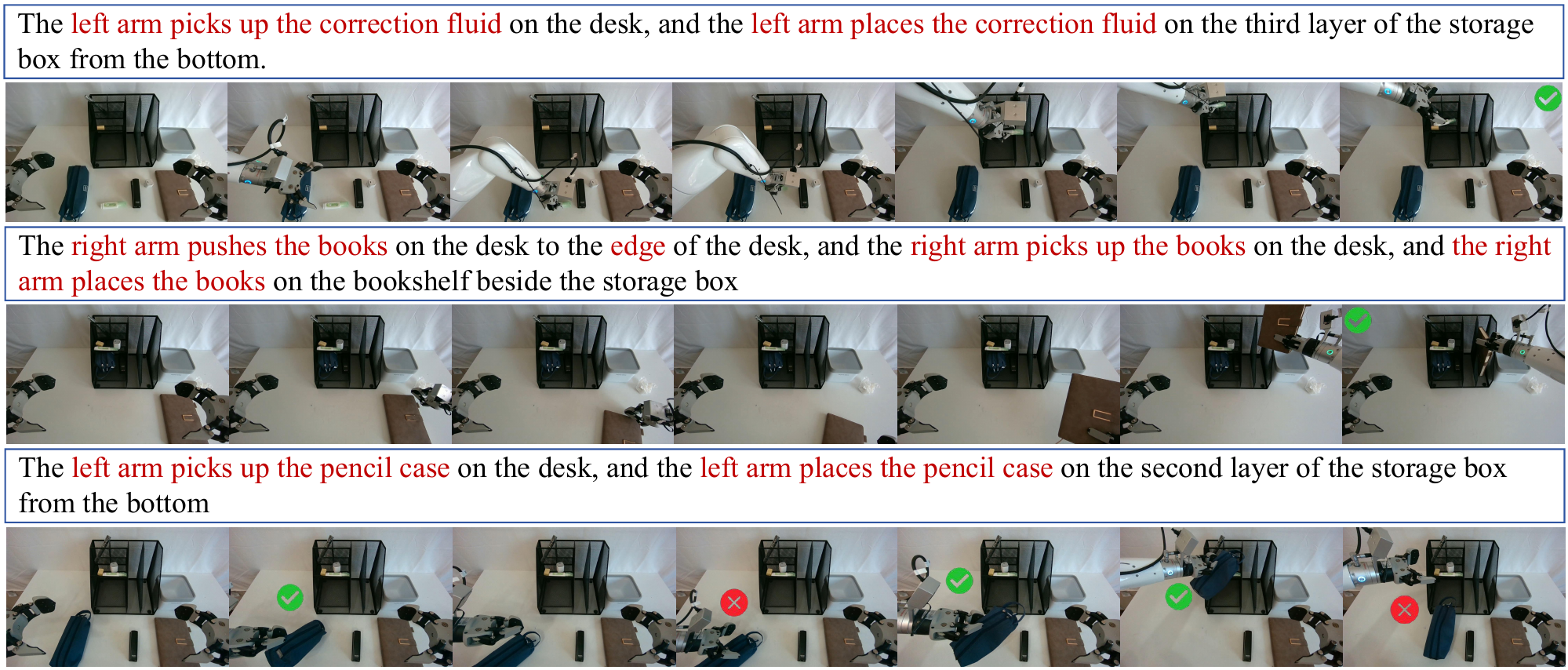}
    \caption{Visualization results of the real-robot policy executions. The top two rows show successful task executions, while the bottom row presents examples of failed executions.}
    \label{fig:real_policy}
\end{figure*}

\begin{table*}[t]
\centering
\small
\caption{Real-robot success rate (\%) on the eight desktop subtasks. S: per-task policies (single-task); M: one policy post-trained on the mixture of all eight subtask splits and evaluated per subtask (multi-task). Best per column in \textbf{bold}.}
\label{tab:realbot}
\setlength{\tabcolsep}{4pt}
\begin{tabular}{lcccccccccccc}
\toprule
& \multicolumn{2}{c}{Baseline} & \multicolumn{2}{c}{LAPA} & \multicolumn{2}{c}{Moto} & \multicolumn{2}{c}{UniVLA} & \multicolumn{2}{c}{DreamDojo} & \multicolumn{2}{c}{\textbf{GDLAM}} \\
\cmidrule(lr){2-3}\cmidrule(lr){4-5}\cmidrule(lr){6-7}\cmidrule(lr){8-9}\cmidrule(lr){10-11}\cmidrule(lr){12-13}
Subtask & S & M & S & M & S & M & S & M & S & M & S & M \\
\midrule
Pen (L)         & 20 & 5  & 20 & 0  & 20 & 5  & 25 & 5  & 25 & 5  & \textbf{30} & \textbf{10} \\
Eraser (R)      & 25 & 5  & 25 & 5  & 30 & 5  & 30 & 10 & 35 & 10 & \textbf{40} & \textbf{15} \\
Correction (L)  & 15 & 0  & 15 & 0  & 15 & 0  & 20 & 5  & 20 & 5  & \textbf{25} & \textbf{10} \\
Charger (R)     & 60 & 35 & 55 & 35 & 60 & 40 & 60 & 45 & 60 & 45 & \textbf{65} & \textbf{50} \\
Pencil case (L) & 60 & 45 & 55 & 45 & 60 & 50 & 60 & 55 & 65 & 55 & \textbf{65} & \textbf{60} \\
Stapler (R)     & 70 & 45 & 70 & 50 & 70 & 55 & 75 & 55 & 75 & 60 & \textbf{80} & \textbf{60} \\
Books (R)       & 35 & 30 & 35 & 30 & 35 & 35 & 40 & 35 & 40 & 40 & \textbf{45} & \textbf{40} \\
Trash (R)       & 60 & 70 & 60 & 75 & 65 & 80 & 70 & 80 & 70 & 85 & \textbf{75} & \textbf{85} \\
\midrule
\textbf{Average} & 43.1 & 29.4 & 41.9 & 30.0 & 44.4 & 33.8 & 47.5 & 36.3 & 48.8 & 38.1 & \textbf{53.1} & \textbf{41.3} \\
\bottomrule
\end{tabular}
\end{table*}

\subsection{G1. Real-Robot Deployment and Evaluation}
\label{app:realbot}

To probe whether GDLAM's group-disentangled code transfers beyond simulation, we take the learned latent action representation as the pretraining backbone for a downstream policy learner, JoyAI-RA-0.1 \citep{zhang2026joyai}, and fine-tune the resulting bimanual policy on a physical robot performing a suite of desktop manipulation tasks. Each latent action model (including GDLAM and the single-code baselines) is coupled with the same JoyAI-RA-0.1 policy learner, so that the comparison isolates the quality of the latent action code.

\paragraph{Platform and tasks.} We use an AgiBot G1 \citep{bu2025agibot} humanoid equipped with two arms and parallel grippers, teleoperated to collect a real-robot dataset (\textbf{g1\_all\_realbot}) in a tabletop (desk) scene. The benchmark comprises eight desktop subtasks that require picking a target object and placing it into a designated compartment of a mesh storage box (or a trash can), collected as eight equally weighted (sampling ratio $1.0$) LeRobot-format splits: \textbf{g1\_pen}, \textbf{g1\_eraser}, \textbf{g1\_correction} (correction fluid), \textbf{g1\_charger}, \textbf{g1\_pencil} (pencil case), \textbf{g1\_stapler}, \textbf{g1\_books}, and \textbf{g1\_trash}. Each subtask is specified by a natural-language instruction naming the arm, the source object, and the target location (e.g., \emph{``The left arm picks up the pen on the desk and places it into the pen holder at the top of the storage box''}), so the policy must ground both the manipulated object and its destination compartment. Fig.~\ref{fig:realbot} shows the physical setup and a representative rollout.

\paragraph{Observation and action spaces.} Every trajectory provides three synchronized RGB streams---a head-mounted camera and left/right wrist cameras---each resize-padded to $224{\times}224$ to match the frozen-VFM tokenizer. Proprioceptive state is min--max normalized to $[-1,1]$ and zero-padded to a fixed $64$-dimensional vector for a uniform interface across embodiments. Actions are the bimanual joint targets with $\mathrm{action\_dim}{=}32$, laid out as left arm ($7$) / left gripper / right arm ($7$) / right gripper; gripper channels are kept as continuous values and thresholded on the robot side.

\paragraph{Training configuration.} The real-robot policy is obtained by composing three stages: (i) the JoyAI-RA-0.1 base weights provide the initial policy backbone; (ii) \textbf{Latent action pretraining} follows the same protocol as Appendix G, where latent actions are extracted by different methods and used for VLA pretraining; and (iii) the resulting policy is \textbf{post-trained} on \textbf{g1\_all\_realbot} with real action supervision and the three-camera input above. Optimization uses AdamW ($\beta{=}(0.9,0.95)$) with a base learning rate of $1{\times}10^{-4}$ under a cosine-with-min-lr schedule, $5000$ warmup steps, per-device batch size $16$, and up to $3{\times}10^{5}$ steps. We enable mixed-precision training with gradient checkpointing and clip gradients at $\mathrm{max\_grad\_norm}{=}1.0$. Training runs on a multi-node cluster with $8$ GPUs per node.

\paragraph{Evaluation protocol.} Our reference Baseline starts from the same JoyAI-RA-0.1 base weights but skips the latent action pretraining stage, post-training directly on \textbf{g1\_all\_realbot} with no latent action representation. Against it we compare four latent action pretraining schemes---LAPA, Moto, UniVLA, and DreamDojo---and our group-disentangled GDLAM, each used to initialize the identical policy. For every method we report per-subtask success rate ($20$ trials per subtask) under two regimes: single-task, where a separate policy is post-trained and evaluated per subtask, and multi-task, where the eight subtask splits are mixed together into a single post-training set to obtain one policy, which is then evaluated on each subtask separately.

\paragraph{Results.} Table~\ref{tab:realbot} reports the full per-subtask breakdown and Fig.~\ref{fig:real_policy} presents qualitative visualizations of successful and failed executions by the real-robot policy. All latent action pretraining schemes improve over the Baseline, confirming that latent action pretraining transfers to real hardware. Among them, GDLAM attains the best average success in both regimes ($53.1\%$ single-task, $41.3\%$ multi-task), for an overall gain of about $+10$ points over the Baseline ($43.1\%/29.4\%$) and a consistent edge over the strongest prior pretraining scheme (DreamDojo). The improvements concentrate on the fine-grained, small-object tasks (pen, eraser, correction fluid), where an unstructured latent code struggles to separate the reach--grasp--insert phases; GDLAM's group-disentangled code preserves these controllable factors and transfers more reliably. The margin is largest in the harder multi-task regime, where a single policy is post-trained on the mixture of all eight subtask splits: the Baseline collapses to $0\%$ on the tightest insertions (pen, correction fluid), whereas GDLAM retains non-trivial success across every subtask.

\section{H. Causal-Intervention Protocols}
\label{app:causal}
Correlational metrics such as Modularity, MIG, and DCI can be satisfied by a code whose units are merely decorrelated on the evaluation distribution, without that separation holding under a genuine intervention. This appendix specifies the three protocols we use to test disentanglement causally, how each maps onto the quantities reported in Table~2 and Fig.~1(a), and the hyperparameters used to run them. Unless stated otherwise, every protocol is run on 500 held-out clip pairs from Bridge V2 and LIBERO, disjoint from LAM pretraining and from the MI/probing splits in Appendix~B, and averaged over four seeds.

\begin{table}[t]
\centering
\caption{Leave-one-out ablation of the four disentanglement objectives (one term zeroed at a time; all else at default).}
\label{tab:loo}
\begin{tabular}{lcc}
\toprule
Variant & DCI$\uparrow$ & MI(KSG)$\uparrow$ \\
\midrule
Full model (default)        & \textbf{0.88} & \textbf{1.24} \\
\ \ w/o $\mathcal{L}_{\mathrm{mx}}$   & 0.61 & 1.21 \\
\ \ w/o $\mathcal{L}_{\mathrm{ge}}$   & 0.74 & 1.22 \\
\ \ w/o $\mathcal{L}_{\mathrm{gs}}$   & 0.80 & 1.23 \\
\ \ w/o $\mathcal{L}_{\mathrm{orth}}$ & 0.83 & 1.23 \\
\bottomrule
\end{tabular}
\end{table}

\begin{table}[t]
\centering
\caption{Ablation of architectural mechanisms and the anti-leak augmentation.}
\label{tab:arch-abl}
\begin{tabular}{lcc}
\toprule
Variant & DCI$\uparrow$ & MI(KSG)$\uparrow$ \\
\midrule
Full model (default)         & \textbf{0.88} & \textbf{1.24} \\
\ \ shared (non-isolated) bottleneck & 0.64 & 1.20 \\
\ \ uniform gate (no spatial gate)   & 0.71 & 1.22 \\
\ \ w/o anti-leak augmentation       & 0.82 & 1.06 \\
\bottomrule
\end{tabular}
\end{table}

\begin{table}[ht]
\centering
\caption{Effect of the number of groups $N$ on disentanglement, information content, and training cost.}
\label{tab:group-count}
\begin{tabular}{lccc}
\toprule
$N$ & DCI$\uparrow$ & MI (KSG)$\uparrow$ & Rel.\ step time \\
\midrule
1 (Single-code) & 0.48 & 0.94 & 0.81$\times$ \\
2               & 0.69 & 1.10 & 0.89$\times$ \\
4 (default)     & \textbf{0.88} & \textbf{1.24} & 1.00$\times$ \\
8               & 0.85 & 1.22 & 1.34$\times$ \\
\bottomrule
\end{tabular}
\end{table}

\begin{table}[ht]
\centering
\caption{Sensitivity to disentanglement-weight scaling and to the static-group designation. ``Leakage'' is the fraction of dynamic-group gate mass falling on the static background ($\downarrow$).}
\label{tab:sens}
\begin{tabular}{lccc}
\toprule
Variant & DCI$\uparrow$ & MI(KSG)$\uparrow$ & Leakage$\downarrow$ \\
\midrule
Full model (default)        & \textbf{0.88} & \textbf{1.24} & \textbf{0.09} \\
\ \ all weights $\times0.5$  & 0.85 & 1.23 & 0.12 \\
\ \ all weights $\times2.0$  & 0.86 & 1.19 & 0.08 \\
\ \ no static-group & 0.86 & 1.23 & 0.19 \\
\bottomrule
\end{tabular}
\end{table}

\paragraph{(1) Single-group intervention $\to$ response matrix.} For a held-out transition $(x_t,x_{t+1})$, we encode the grouped code $\{a_t^{(g)}\}_{g=1}^N$, then perturb one group at a time: $a_t^{(j)} \leftarrow a_t^{(j)} + \epsilon$ with $\epsilon\sim\mathcal{N}(0,\sigma^2 I)$, $\sigma$ set to the empirical per-group latent standard deviation (we also repeat with full resampling, $a_t^{(j)}\leftarrow \tilde a^{(j)}$ drawn from the group's marginal, and find both variants agree within noise). We decode the perturbed code to a predicted future frame and re-encode it to obtain the updated code $\{a_t'^{(i)}\}$. The response of group $i$ to an intervention on group $j$ is the gate-weighted, normalized change
\begin{equation}
M_{ij} = \frac{\big\| \mathrm{Gate}_t^{(i)} \odot \big(D_\theta(z_t, a_t') - D_\theta(z_t, a_t)\big) \big\|_2}{\big\| \mathrm{Gate}_t^{(i)} \odot D_\theta(z_t, a_t) \big\|_2 + \epsilon_0},
\end{equation}
i.e., the change in the decoded feature map restricted to group $i$'s own spatial gate, normalized by that region's baseline magnitude. Averaging $M$ over the evaluation set gives the response matrix in Fig.~1(a); from it we read off the diagonal response $\mathrm{Diag}=\frac{1}{N}\sum_j M_{jj}$ and the off-diagonal leakage $\mathrm{Leakage}=\frac{1}{N(N-1)}\sum_{i\neq j} M_{ij}$ reported in Table~2. For the monolithic baselines, which expose no gate, we substitute a fixed half-image split of the code's decoder attention map for $\mathrm{Gate}^{(i)}$, applied identically to both halves.

\paragraph{(2) Cross-group probing $\to$ Cross-$R^2$.} To test whether a group's information is redundantly recoverable from the rest of the code, we fit, for each group $k$, a two-layer MLP regressor $g_{-k}$ that predicts $a_t^{(k)}$ from the concatenation of all remaining groups $\{a_t^{(j)}\}_{j\neq k}$, trained on a held-out training split and evaluated on a disjoint test split. We report the coefficient of determination $R^2_k$ of this regressor, and the Cross-$R^2$ column of Table~2 is the average $\frac{1}{N}\sum_k R^2_k$; a low value means each group carries information the others cannot reconstruct, which is the complement of the leakage statistic above (leakage measures whether an intervention on one group spills into another's output region, while Cross-$R^2$ measures whether one group's content is already implicit in the others, even without any intervention).

\paragraph{(3) Group-swap sweep.} Protocol (1) perturbs a group with synthetic noise or resampling; as a robustness check that our conclusions are not an artifact of that synthetic perturbation, we repeat a swap sweep using only real content: for every ordered pair of held-out clips and every group index $k$, we splice in group $k$ from the second clip and accumulate the resulting $M_{ik}$ values into a second response matrix. Across our evaluation set this matrix agrees with the one from protocol (1) to within $0.03$ on every entry for GDLAM, confirming the diagonal structure in Fig.~1(a) is not specific to how the intervention is generated.

\begin{table*}[t]
\centering
\caption{Compute comparison. Params: total; PeakMem: peak forward activation memory; Fwd: mean forward latency.}
\label{tab:cost-compare}
\begin{tabular}{lccccc}
\toprule
Model & Params (M) & Trainable (M) & PeakMem (GB) & GFLOPs & Fwd (ms) \\
\midrule
UniVLA  & 287 & 201 & 2.0 & 132 & 107 \\
LAPA       & 344 & 344 & 2.8 & 43  & 197 \\
Moto      & 439 & 121 &  2.2 & 148 & 69  \\
MVP-LAM        & 287 & 201 & 5.6 & 308 & 90  \\
DreamDojo    & 710 & 710 & 5.6 & 225 & 117 \\
\textbf{GDLAM (ours)} & 392 & 285 & 3.1 & 168 & 101 \\
\bottomrule
\end{tabular}
\end{table*}

\section{I. Ablations: Isolating Every Design Choice}
\label{app:ablation}
The main paper attributes GDLAM's disentanglement to a specific set of design choices: isolated per-group bottlenecks, spatially gated routing, four geometric objectives ($\mathcal{L}_{\mathrm{mx}},\mathcal{L}_{\mathrm{gs}},\mathcal{L}_{\mathrm{ge}},\mathcal{L}_{\mathrm{orth}}$), an anti-leak augmentation, and the choice of $N$. We ablate each component in turn, holding all else fixed at the default configuration (Table~\ref{tab:hparam}), and report two complementary axes so that no single metric can hide a regression: DCI ($\uparrow$, disentanglement) and MI(KSG) ($\uparrow$, action-informativeness). All numbers are means over four seeds; the default full model is repeated in every table as the reference row. This design lets a reader verify that every component pays for itself on at least one axis without silently costing another.

\paragraph{(1) Leave-one-out on the four objectives.} Table~\ref{tab:loo} removes one geometric term at a time. Removing mutual exclusivity ($\mathcal{L}_{\mathrm{mx}}$) is the single most damaging change to disentanglement---DCI collapses from $0.88$ to $0.61$---confirming it is the primary force separating group directions. Dropping gate sparsity ($\mathcal{L}_{\mathrm{ge}}$) leaves DCI moderately reduced ($0.88\!\to\!0.74$), because without a compact mask a group's influence bleeds across the image even when its code direction is distinct---a spatial-locality failure that DCI alone would under-report. Removing group sparsity ($\mathcal{L}_{\mathrm{gs}}$) mainly hurts the quiet-unless-active property, mildly lowering DCI. Removing static--dynamic orthogonality ($\mathcal{L}_{\mathrm{orth}}$) has the smallest headline effect on DCI but measurably worsens the specific static/dynamic confusion (appearance drift leaking into motion groups), which we quantify separately in ablation (4).

\paragraph{(2) Architectural components.} Table~\ref{tab:arch-abl} ablates the two structural mechanisms. Replacing the isolated per-group bottleneck with a single shared bottleneck (splitting one code post-hoc) drops DCI to $0.64$: without a per-group rate--distortion cost, groups silently share capacity, so factorization degrades even though total MI is preserved. Replacing the learned spatial gate with a uniform (all-ones) gate---i.e.\ every group influences every token equally---leaves code-direction disentanglement partly intact (DCI $0.71$) but degrades the factorization measured by DCI, mirroring the $\mathcal{L}_{\mathrm{ge}}$ result and confirming that spatial specialization and directional separation are complementary, not redundant. Removing the anti-leak augmentation lets the predictor copy static appearance through the residual path, hollowing out the code (MI drops to $1.06$), a shortcut that DCI alone would not flag as strongly as MI does.

\paragraph{(3) Number of groups.} Table~\ref{tab:group-count} reports the group-count sweep $N\in\{1,2,4,8\}$, with wall-clock overhead relative to $N{=}4$; here $N{=}1$ acts as a single-bottleneck ablation of our own architecture (GDLAM stripped of its grouping). DCI peaks at $N{=}4$ ($0.88$) and declines at $N{=}8$ ($0.85$) once groups outnumber the distinct causes in the data; MI(KSG) is nearly flat, confirming $N$ reorganizes rather than adds information. Increasing $N$ to $8$ raises per-step cost by $\approx34\%$ (extra per-group cross-attention projections), while $N{=}1,2$ are cheaper but under-factorize the code. $N{=}4$ is the accuracy/compute sweet spot.

\paragraph{(4) Static group and objective-weight sensitivity.} Table~\ref{tab:sens} varies each disentanglement weight by $\pm2\times$ around its default and toggles the designated static group. GDLAM is robust to weight perturbations: DCI stays within $\pm0.03$ across all $\pm2\times$ settings, indicating we are not sitting on a sharp optimum that only careful tuning finds. Removing the static-group designation (all groups treated as dynamic) leaves overall DCI nearly unchanged ($0.86$) but roughly doubles static/dynamic leakage---appearance drift measurably contaminating motion groups, quantified as the off-block gate mass falling on the static background---which is precisely the targeted failure $\mathcal{L}_{\mathrm{orth}}$ addresses and which a single aggregate DCI number would mask.

\section{J. Compute and Training Cost}
\label{app:cost}
To ensure a fair comparison of computational cost between GDLAM and existing LAMs, we evaluate all methods under a unified profiling protocol. All measurements are conducted on a single GPU using a pair of (224$\times$224) RGB frames as input (batch size (1)) in FP32 precision. We report the parameter count of the full model, the peak activation memory allocated during a forward pass, the average inference latency measured after warm-up and synchronization, and the forward-pass FLOPs estimated from the computation graph. GDLAM is designed to balance representational capacity with computational efficiency. Its perceptual tokenizer is a frozen VFM, introducing no optimizer state during training, while the trainable components consist only of a lightweight delta encoder, grouped action encoder, and grouped predictor with depths of 6, 6, and 12 transformer blocks, respectively. Moreover, the grouped cross-attention module is implemented as a single batched attention operation over all latent groups, rather than separate attention modules for each group, preventing computational cost from scaling linearly with the number of groups. As summarized in Table~\ref{tab:cost-compare}, GDLAM occupies a moderate position among representative LAMs in terms of parameter count, memory footprint, FLOPs, and inference latency, achieving runtime efficiency comparable to mainstream VFM-tokenized latent action models. The proposed group-disentanglement mechanism incurs only modest computational overhead while substantially improving representation disentanglement and downstream controllability and generalization, demonstrating a favorable trade-off between performance and efficiency.

\end{document}